\RequirePackage{fix-cm}
\documentclass[twocolumn]{svjour3}          
\smartqed  
\usepackage{graphicx}
\usepackage{multirow}
\usepackage[lined,ruled,linesnumbered]{algorithm2e}
\usepackage{url}            
\usepackage{booktabs}       
\usepackage{amsfonts}       
\usepackage{nicefrac}       
\usepackage{microtype}      
\usepackage{graphicx}
\usepackage{subfigure}
\usepackage{amsmath,amssymb}
\usepackage{caption}
\usepackage{color}
\usepackage{url}
\usepackage[breaklinks=true]{hyperref}
\usepackage{textcomp}
\usepackage{bbding}
\usepackage{pifont}

\begin{document}
\hyphenpenalty=1000
\title{Large-scale Bisample Learning on ID Versus Spot Face Recognition}


\author{Xiangyu Zhu$^{*}$ \and Hao Liu$^{*}$ \and Zhen Lei \and Hailin Shi \and Fan Yang \and Dong Yi \and Guojun Qi \and Stan Z. Li
}


\institute{Xiangyu Zhu, Hao Liu, Zhen Lei, Hailin Shi and Stan Z. Li \at
              Center for Biometrics and Security Research \& National Laboratory of Pattern Recognition, Institute of Automation,
              Chinese Academy of Sciences. \\
              \email{\{xiangyu.zhu,hao.liu2016,zlei,hailin.shi,szli\}@nlpr.ia.ac.cn}           
              \and
           Fan Yang \at
              College of Software, Beihang University.\\
              \email{fanyang@buaa.edu.cn}
              \and
           Dong Yi \at
              DAMO Academy, Alibaba Group.\\
              \email{yidong.yd@alibaba-inc.com}
              \and
           Guojun Qi \at
              HUAWEI Cloud, USA.\\
              \email{guojunq@gmail.com}
              \and
              $^{*}$denotes equal contribution
}

\date{Received: date / Accepted: date}

\maketitle

\begin{abstract}

In real-world face recognition applications, there is a tremendous amount of data with two images for each person. One is an ID photo for face enrollment, and the other is a probe photo captured on spot. Most existing methods are designed for training data with limited breadth (a relatively small number of classes) and sufficient depth (many samples for each class). They would meet great challenges on ID versus Spot (IvS) data, including the under-represented intra-class variations and an excessive demand on computing devices. In this paper, we propose a deep learning based large-scale bisample learning (LBL) method for IvS face recognition. To tackle the bisample problem with only two samples for each class, a classification-verification-classification (CVC) training strategy is proposed to progressively enhance the IvS performance. Besides, a dominant prototype softmax (DP-softmax) is incorporated to make the deep learning scalable on large-scale classes. We conduct LBL on a IvS face dataset with more than two million identities. Experimental results show the proposed method achieves superior performance to previous ones, validating the effectiveness of LBL on IvS face recognition.

\end{abstract}

\section{Introduction}\label{sec-introduction}

\begin{figure}
  \centering
  \includegraphics[width=0.5\textwidth]{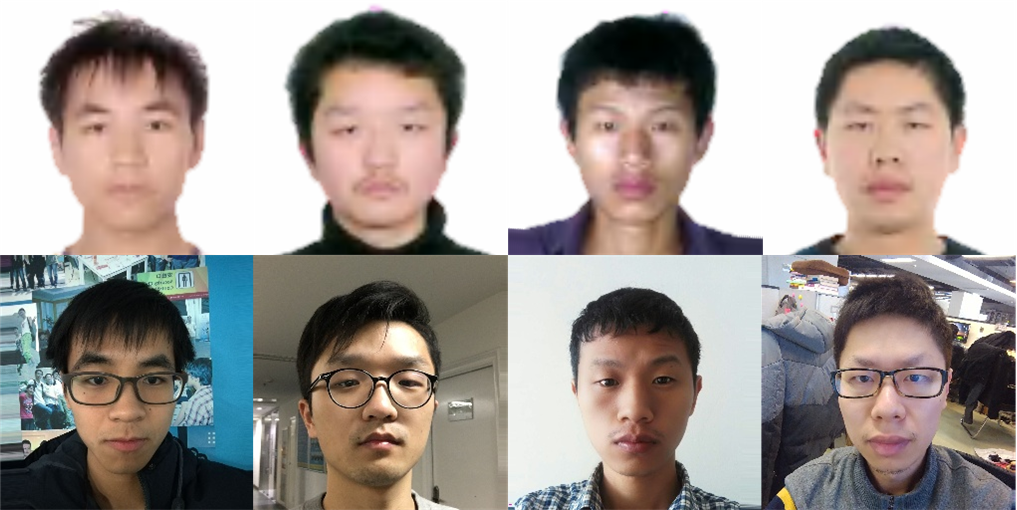}
  \caption{The ID versus Spot (IvS) data, each identity has one ID photo and one spot photo.}
  \label{fig-ivs-demo}
\end{figure}

Face recognition has witnessed dramatic improvements in recent years, primarily due to the advances in network architectures~\cite{Krizhevsky2012ImageNet,szegedy2014going,simonyan2014very,szegedy2016rethinking,he2016deep}, training strategies~\cite{Taigman-CVPR-2013,sun2014deep,schroff2015facenet,Smirnov2017Doppelganger,huang2016local,Chen2017Beyond} and a large amount of face data~\cite{yi2014learning,nech2017level,guo2016ms,Cao2017VGGFace2}. Recent methods mainly focus on face recognition in the wild, where the training datasets are collected from internet by web searching engines~\cite{yi2014learning} or electronic album applications~\cite{nech2017level}. Most of wild datasets like CASIA-Webface~\cite{yi2014learning}, Ms-Celeb-1M~\cite{guo2016ms} and VGG2~\cite{Cao2017VGGFace2} are well-posed, where they have a limited number of classes (less than $100,000$) and adequate samples per class (more than $20$). However, this is not the case in many real-world face data, like the ID versus Spot (IvS) face recognition, which aims to match unconstrained spot photos with constrained ID photos, see Fig.~\ref{fig-ivs-demo} for example.
Compared with wild datasets, IvS datasets present threefold challenges below.

\begin{enumerate}
\item \textbf{Heterogeneity}: ID and spot photos are taken in different environments. The ID photos are taken in constrained environments with clean background, in frontal pose, normal illumination and neutral expression. The spot photos are taken in unconstrained environments. There are pose, lighting, expression and occlusion (e.g., glasses, haircut, scarf etc.) variations. Moreover, there may be a large age gap between ID and spot photos since ID photos are updated every $10-20$ years. This heterogeneity increases the difficulty of IvS face recognition.
\item \textbf{Bisample Data}: Usually, IvS training data is collected by face authentication systems. When a user passes the authentication system, a pair of his photos will be recorded, one ID photo from his ID card and the other spot photo taken online. As a result, there are only two samples available for each subject. The intra-variations of classes are not well represented, making the discriminative training on bisample data a more challenging problem. 
\item \textbf{Large-scale Classes}: IvS data is collected by practical systems, where there can be as many as million or even hundreds of million identities. How to perform deep learning on such massive classes with limited GPU devices is worth studying.
\end{enumerate}

The above three characteristics pose great challenges for IvS face recognition. In real-world applications, the high recognition rate at low false acceptance rate is demanded. To this end, the large margins between inter-class samples and the compactness of intra-class samples in the feature space are necessary. However, since there are only two samples for each subject, it is difficult to describe the intra-variations in the training phase so that the derived feature space would not be discriminative enough. In addition, there is a huge number of classes. It is a great challenge to explore the discriminative information among these classes with limited GPU devices. Taking deep learning with softmax as an example, there need to be millions of prototypes in the GPU memory, which is infeasible for most of computing devices.

In this paper, we cast the deep learning on IvS data as a \textbf{L}arge-scale \textbf{B}isample \textbf{L}earning (LBL) problem, where the training data has a huge number of classes and each class has only one positive pair. To enhance existing training strategies to handle the LBL problem, two challenges must be resolved: The weak intra-variations caused by bisample data and the model training scalability caused by large-scale classes. To deal with weak intra-variations, we propose a progressive model transferring method, named \textbf{C}lassification-\textbf{V}erification-\textbf{C}lassification (CVC). We pre-train a model on web-collected data by classification and finetune it on IvS data by verification to get a good initialization. Then we perform large-scale classification to obtain the final IvS model.

To improve scalability for model training, we adopt a prototype selection strategy in the last stage of CVC to scale up softmax-like losses to any number of classes. Specifically, we observe that the gradients of softmax are dominated by a small fraction of classes and the dominant classes can be effectively identified by the class proximities. Based on this, we build a dominant queue for each class to record its similar classes, from which we can select the most dominant classes to participate in the classification. The new softmax can perform effective training with only $0.15\%$ classes, significantly reducing the demand for computing devices.

We evaluate our method on a real-world IvS dataset and show it reaches the state-of-the-art performance with limited computing devices (4 TITANX GPU). Besides, we release a Public-IvS dataset of $1262$ identities for open evaluation \footnote{http://www.cbsr.ia.ac.cn/users/xiangyuzhu/} . Moreover, to make our work reproducible, we devise a new protocol Megaface-bisample to mimic the large-scale bisample learning task. To our knowledge, it is the first investigation into training deep neural networks on large-scale bisample face data.

\section{Related Works}\label{sec-related-works}
In this section, we review the deep learning based face recognition and discuss two related problems about the LBL task: (1) Learning with insufficient data and (2) Large-scale classification.

\subsection{Deep Learning based Face Recognition}
Recently there are two schemes to train deep models for face recognition: classification and verification. The classification scheme considers each identity as a unique category and classifies each sample into one of the classes. During testing, the classification layer is removed and the top-level feature is regarded as the face representation~\cite{Sun-CVPR-2014}. The most popular loss is softmax~\cite{Sun-CVPR-2014,Taigman-CVPR-2013,taigman2014web}. Based on that, the center loss~\cite{wen2016discriminative} proposes to learn the class-specific feature centers to make features more compact in the embedding space. The L2-softmax~\cite{ranjan2017l2} adds a L2-constraint on features to promote the under-represented classes. The normface~\cite{wang2017normface} normalizes both features and prototypes to make the training and testing phases closer. Recently, enhancing margins between different classes is found to be effective in improving feature discrimination, including large-margin softmax~\cite{liu2016large}, A-softmax~\cite{liu2017sphereface}, GA-softmax~\cite{liu2017deephyper} and AM-softmax~\cite{wang2018additive}. Benefiting from the prototypes in the classification layer, the scheme can distinguish a sample from all the other classes, leading to fast convergence and good generalization ability~\cite{wang2017normface}.

On the other hand, the verification scheme optimizes distances between samples. Within a mini-batch, the contrastive loss~\cite{sun2014deep} optimizes pairwise distances in the feature space to reduce intra-class distances and enlarge inter-class distances. The triplet loss~\cite{schroff2015facenet} makes up a triplet consisting of an anchor, a positive sample and a negative sample. The loss aims to separate the positive pair from the negative pair by a distance margin. The lifted structured loss~\cite{oh2016deep} considers all the pairwise distances within the mini-batch and select the best positives and negatives. The N-pairs loss~\cite{sohn2016improved} optimizes each positive pair against all the related negative pairs following a local softmax formulation. Besides, hard negative mining is widely adopted to remove the easy negative pairs to ensure fast convergence~\cite{schroff2015facenet}. More recently, \cite{zhao2018princi} presents a GAN-based method to deliberately generate hard triplet samples to improve the efficiency and effectiveness in training triplet losses. The performance of the verification scheme depends on the number of pairs generated in one mini-batch~\cite{oh2016deep}, which is determined by the batch size. However, increasing batch size, meaning that expanding GPU memory, is very expensive. To reduce the cost of GPU memory, smart sampling~\cite{kumar2017smart} selects valuable pairs in the data layer instead of the feature layer. The method memorizes the pairs having large losses and selects them with higher probabilities afterwards~\cite{kumar2017smart,Smirnov2017Doppelganger,wang2017train}.

Most contemporary face recognition methods are based on wild datasets, e.g., CASIA-Webface~\cite{yi2014learning}, Ms-Celeb-1M~\cite{guo2016ms}, MF2~\cite{nech2017level} and VGG2~\cite{Cao2017VGGFace2}. These well-posed datasets have a limited number of identities and sufficient samples per identity. However, this is not the case in IvS datasets. Table~\ref{tab-dataset} gives a brief comparison between wild and IvS datasets. Our CASIA-IvS has more than $2$ million identities but only two samples per identity, on which existing well-studied methods cannot work well any more. Exploring IvS-specific training strategies is necessary.

\begin{table*}[!htb]
 \tabcolsep 7pt
  \begin{center}
  \begin{tabular}{ c  c  c  c  c  c}
    \hline
    \multirow{2}{*}{Dataset} & \multirow{2}{*}{Identities} & \multirow{2}{*}{Samples/ID} & \multirow{2}{*}{Scenarios}  & \multirow{2}{*}{Descriptions} \\
    &  &  &  &      \\
    \hline
    \hline
    CASIA-Webface~\cite{yi2014learning} & $10,575$ & $46.7$ & wild & Celebrity photos by web-searching  \\
    \hline
    Ms-Celeb-1M~\cite{guo2016ms} & $98,685/79,077$ & $50.7/63.8$ & wild & Celebrity photos by web-searching  \\
    \hline
    MF2~\cite{nech2017level} & $657,559/90,399$ & $6.8/24.7$ & wild & User photos of electronic album\\
    \hline
    VGG2~\cite{Cao2017VGGFace2} & $9,131$ & $362.6$ & wild & Celebrity photos by web-searching \\
    \hline
    \hline
    CASIA-IvS & $2,578,178$ & $2$ & IvS & ID and spot photos of the masses \\
    \hline
  \end{tabular}
  \end{center}
  \caption{Description of face recognition datasets. We clean Ms-Celeb-1M and MF2 due to their low purities~\cite{wu2015light}, and cut the identities whose samples are smaller than $10$ to balance the long tail distribution~\cite{zhang2017range}. The numbers after $/$ indicate the information after cleaning.}
  \label{tab-dataset}
\end{table*}

\subsection{Learning with Insufficient Data}
\textbf{Low-shot learning} intends to recognize new classes by few samples~\cite{feifei2006one-shot}. Generally, low-shot learning transfers the knowledge from a well-posed source domain to the low-shot target domain. Siamese net~\cite{koch2015siamese} trains a siamese CNN by same-or-different classification on the source domain and extracts the deep features for nearest neighbour matching in the target domain. MANN~\cite{santoro2016one-shot,weston2014memory,vinyals2016matching} memorizes the features of examples in the source domain to help predict the under-labeled classes. Model regression~\cite{Wang2016Learning,bertinetto2016learning} directly transfers the neural network weights across domains. The L2-regularization on features~\cite{guo2017one,hariharan2016low,chen2016analysis} can prevent the network from ignoring low-shot classes. Besides, virtual sample generation~\cite{hariharan2016low,choe2017face} and semi-supervised samples~\cite{xu2016few} are found effective in promoting low-shot classes. Although both low-shot learning and bisample learning intend to learn a concept with insufficient samples, they differ in that low-shot learning is close-set classification but bisample learning is open-set classification where the testing samples definitely belong to unseen classes.

\textbf{Long-tail problem} refers to the situation that only a limited number of classes appear frequently, while most of the others remain far less existing. Deep models trained on long-tailed data tend to ignore the classes in the tail. To resolve the problem, \cite{Yang2014Context} retrieves more samples from the tail classes. \cite{Ouyang2016Factors} makes samples uniformly distributed by random sampling. \cite{zhang2017range} proposes a range loss to balance the rich and poor classes, where the largest intra-class distance is reduced and the shortest class-center distance is enlarged.

\subsection{Large-scale Classification}
Large-scale classification aims to perform classification on a vast number of classes, where the class number reaches millions or tens of millions. This task presents a great problem for deep learning: the common softmax loss can not be adopted due to the prohibitive parameter size and computation cost. The Megaface challenge~\cite{nech2017level} proposes four methods for training models on $670$k identities. Model-A trains the network on random $20,000$ identities via softmax. Model-B finetunes Model-A on all the $670$k identities with the triplet loss. Model-C adopts rotating softmax that randomly selects $2,600$ identities every $20$ epoches. After each rotation the parameters in the softmax layer are randomly initialized. Model-D further triplet-finetunes Model-C on all the identities.

Beyond the computer vision, extreme multi-label learning~\cite{hsu2009multi} and noise contrastive estimation~\cite{gutmann2010noise} are related to large-scale classification. \textbf{Extreme Multi-label Learning} learns a classifier to tag a sample with the most relevant label from a large label set~\cite{hsu2009multi}. It faces the same challenge as LBL that training a multi-class classifier is computationally prohibitive when the class number is extremely large. To tackle this problem, the tree based methods~\cite{choromanska2013extreme,prabhu2014fastxml,bengio2003neural} learn a label hierarchy as follows: The root node contains the entire label set and nodes are recursively partitioned until each leaf contains a small number of labels. Finally a base classifier identifies the samples in only one leaf node.
Although tree based methods reduce the class number for each classifier, the prediction error made at top-level cannot be corrected at lower levels due to its cascading architecture~\cite{babbar2017dismec}. On the other hand, the embedding based methods~\cite{bhatia2015sparse,tagami2017annexml,xu2016robust} assume the label matrix~\cite{hsu2009multi}, where each row is a $\{0,1\}$ label vector of a sample, is low rank and the label vectors can be projected onto a low-dimensional linear subspace. As a result, the extreme classification task can be converted to a low-dimensional regression problem. However, the low rank assumption indicates that the samples concentrate on a small number of active classes, which is not the case in IvS data. 
\textbf{Noise Contrastive Estimation (NCE)}~\cite{gutmann2010noise} provides an approximate method to estimate the probabilistic distribution without the normalization constant, which is the major cost in large-scale classification. Its basic idea is training a logistic regression classifier to discriminate samples from data distribution and noise distribution, so that the density estimation is reduced to probabilistic binary classification. Although NCE has been successfully applied in language models~\cite{mnih2012fast,mnih2013learning,vaswani2013decoding}, recent face recognition tasks~\cite{sun2014deep,Sun-CVPR-2014} have shown that promoting the contrast among classes is crucial in training discriminative models. Turning multi-class classification to binary logistic regression may lose inter-class information and get inferior performance.

\section{Large-scale Bisample Learning}
\begin{figure*}[!htb]
  \centering{}
  \includegraphics[width=0.9\textwidth]{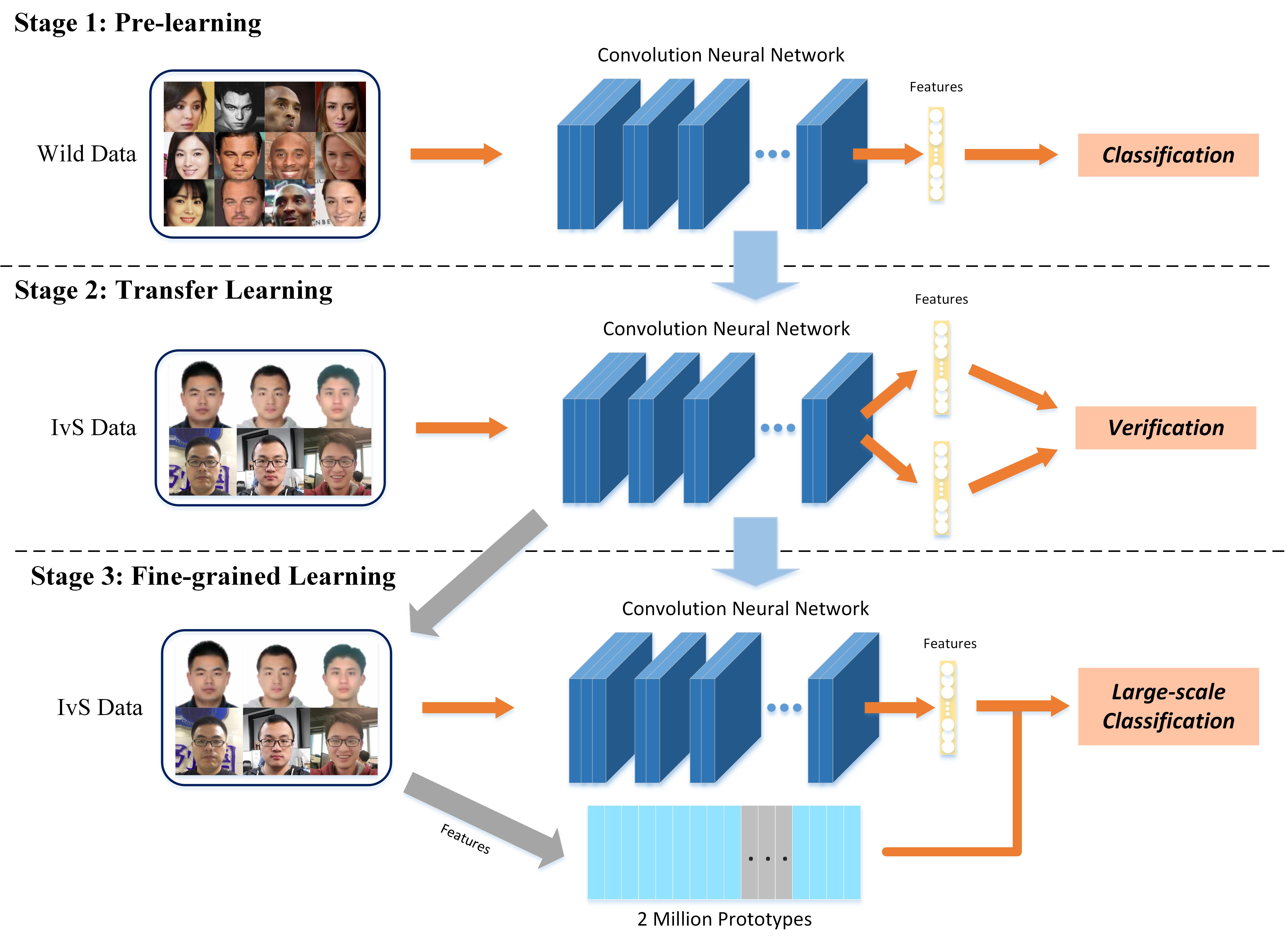}
  \caption{Overview of the large-scale bisample learning (LBL). LBL adopts a classification-verification-classification (CVC) training strategy, which has three stages: The first stage, pre-learning, is training the network from scratch on a wild dataset by a classification loss. The second stage, transfer learning, is finetuning the network on the IvS dataset with a verification loss. The last stage, fine-grained learning, is performing large-scale classification on the IvS dataset with a new dominant prototype softmax. }
  \label{fig-overview}
\end{figure*}

The proposed method contains a complete pipeline for deep learning on large-scale bisample data. We begin by discussing of the classification and the verification schemes, showing how their pros and cons motivate the proposed methods. Then we present the way to train deep neural networks on bisample data. Finally we develop a dominant prototype softmax to perform $2$-million-way classification in a scalable fasion. Fig.~\ref{fig-overview} shows an overview of our method.

\subsection{Problem Formulation and Motivation}
\label{sec:problem_formulation}
Currently there are two schemes for training deep neural networks, i.e., verification and classification. The verification scheme optimizes sample-to-sample distances, such as the contrastive loss~\cite{sun2014deep} and the triplet loss~\cite{schroff2015facenet}. In each iteration, it performs local optimization within a mini-batch by making positive pairs close and negative pairs far away. Besides, the mining strategy~\cite{schroff2015facenet} filters out easy pairs for fast convergence. On the other hand, the classification scheme regards each identity as a unique class and trains the network as a $N$-way classification problem, such as softmax~\cite{Sun-CVPR-2014} and A-softmax\cite{liu2017sphereface}. Compared with the verification scheme, the classification scheme performs global optimization by identifying each sample into one of the $N$ classes.

In this paper, we motivate our method by comparing classification and verification. Interestingly, if we formulate the loss function for a whole mini-batch, we can unify the two schemes in a pair matching and weighting framework. First, the verification scheme extracts features with a neural network and makes pairs between deep features. Taking contrastive loss~\cite{sun2014deep} as an example:
\begin{align}\label{equ-verfication-contrastive}
\emph{L}_{ver}(\mathbf{X}) = -\sum_{j=1}^{M}\sum_{k=j+1}^{M}\emph{NM}(\mathbf{x}_{j}^{\mathrm{T}}\mathbf{x}_{k}, y_{jk})~\mathbf{x}_{j}^{\mathrm{T}}\mathbf{x}_{k}, ~~\mathrm{with}\\ \notag
\emph{NM}(\mathbf{x}_{j}^{\mathrm{T}}\mathbf{x}_{k}, y_{jk}) = \left\{
\begin{aligned}
1 ~~~ &if ~ y_{jk}=1 \\
-1 ~~~  &if ~ y_{jk}=0 ~and~  \mathbf{x}_{j}^{\mathrm{T}}\mathbf{x}_{k} \geq \tau \\
0 ~~~  &if ~ y_{jk}=0 ~and~ \mathbf{x}_{j}^{\mathrm{T}}\mathbf{x}_{k} < \tau \\
\end{aligned}
\right.
\end{align}
where $\mathbf{x}$ is the $D$ dimensional deep feature extracted by the neural network; $\mathbf{X}=[\mathbf{x}_{1},\ldots,\mathbf{x}_{M}]$ are the features in the mini-batch where $M$ is the batch size; $y_{jk}=1$ if $\mathbf{x}_{j}$ and $\mathbf{x}_{k}$ belong to the same class and $y_{jk}=0$ if not; $\emph{NM}(\cdot)$ is the hard negative mining that filters out easy negative pairs with a threshold $\tau$. We can see that the contrastive loss makes pairs within deep features $\mathbf{X}$ and assigns $\{0,1\}$ weights to them.

In contrast, the classification scheme makes pairs between features and prototypes. Taking the softmax loss~\cite{sun2014deep} as an example:
\begin{equation}\label{equ-softmax}
\emph{L}_{cls}(\mathbf{W}, \mathbf{X})=-\sum_{j=1}^{M}\log\big( \frac{e^{\mathbf{w}_{y^{(j)}}^T\mathbf{x}_{j}}}{\sum_{i=1}^{N}e^{\mathbf{w}_{i}^{\mathrm{T}}\mathbf{x}_{j}}} \big),
\end{equation}
where $\mathbf{W}=[\mathbf{w}_{1},\ldots,\mathbf{w}_{N}]$ is the prototype matrix in the softmax layer where $N$ is the number of classes and $y^{(j)}$ is the label of $\mathbf{x}_{j}$. Its derivatives to a prototype $\mathbf{w}_{i}$ and a feature $\mathbf{x}_{j}$ are:
\begin{gather}
\frac{\partial \emph{L}_{cls}}{\partial \mathbf{x}_{j}} = -\sum_{i=1}^{N}(\emph{1}\{y^{(j)}==i\} - p_{ij})\mathbf{w}_{i} \label{equ-softmax-derivative} \\
\frac{\partial \emph{L}_{cls}}{\partial \mathbf{w}_{i}} = -\sum_{j=1}^{M}(\emph{1}\{y^{(j)}==i\} - p_{ij})\mathbf{x}_{j} \notag \\
~\mathrm{with} ~~~~~ p_{ij} = \frac{e^{\mathbf{w}_{i}^T\mathbf{x}_{j}}}{\sum_{k=1}^{N}e^{\mathbf{w}_{k}^{\mathrm{T}}\mathbf{x}_{j}}} \label{equ-softmax-probability} ,
\end{gather}
where $\emph{1}(\cdot)$ is the indicator function which is $1$ when the statement is true and $0$ otherwise, and $p_{ij}$ is the probability that $\mathbf{x}_{j}$ belongs to the $i$th class. Given that network training only concerns the gradients back-propagated, we can construct a dummy softmax loss sharing the same gradients with Equ.~\ref{equ-softmax}:
\begin{align}\label{equ-dummy-softmax}
\emph{L}_{dum}(\mathbf{W}, \mathbf{X}) = -\sum_{j=1}^{M}\sum_{i=1}^{N}~\emph{P}(\tilde{p}_{ij},i,j) \mathbf{w}_{i}^{\mathrm{T}}\mathbf{x}_{j}, ~~\mathrm{with}\\ \notag
\emph{P}(\tilde{p}_{ij},i,j) = \left\{
\begin{aligned}
1-\tilde{p}_{ij} ~~~ &if ~ y^{(j)}=i \\
-\tilde{p}_{ij} ~~~  &if ~  y^{(j)} \neq i
\end{aligned}
\right.
\end{align}
where $\tilde{p}_{ij}$ is computed as $p_{ij}$ in Equ.~\ref{equ-softmax-probability} and considered as a constant. $\emph{L}_{cls}$ and $\emph{L}_{dum}$ are equivalent in network training since they produce the same back-propagated signals. Obviously $\emph{L}_{dum}$ makes pairs between $\mathbf{W}$ and $\mathbf{X}$, and assigns a weight to each pair $(\mathbf{w}_{i}, \mathbf{x}_{j})$ by the probability $p_{ij}$. The negative pairs with higher probabilities and the positive pairs with lower probabilities have larger weights and yield louder signals during training.


Comparing Equ.~\ref{equ-dummy-softmax} and Equ.~\ref{equ-verfication-contrastive}, we can conclude that both classification and verification follow the same pair matching and weighting framework. The only differences lie in the pairing candidates (features with prototypes versus within features) and the weighting methods (soft weight versus hard weight).
Recent works have empirically observed that increasing the number of pairs always delivers faster convergence and better discriminative power, hence the loss functions involving more pairs are preferred. Within a mini-batch with $M$ as the batch size and $N$ as the class number, a classification loss makes $N \times M$ pairs in Equ.~\ref{equ-dummy-softmax} and a verification loss makes $M(M-1)/2$ pairs in Equ.~\ref{equ-verfication-contrastive}. In real implementation with limited GPU memory, $N \gg M$ always holds. For example, when training ResNet64~\cite{liu2016large} with a TITAN-X GPU, the batch size $M$ is about $50$ and the class number $N$ easily reaches tens or even hundreds of thousands. With more orders of magnitude pairs, \textbf{the classification scheme is expected to acquire more discriminative features}, which has been shown in the state-of-the-art methods~\cite{liu2016large,liu2017deephyper,wang2018additive,wang2018cosface}. However, two challenges make classification infeasible on IvS data. First, the classification scheme has difficulty to converge on bisample data due to the weak intra-variations, which is demonstrated in our experiments. Second, the classification scheme suffers from weak scalability to large-scale classes due to the limited GPU memory. Directly performing $2$-million-way classification with two samples per class is infeasible for current optimization methods and computing devices.

In this paper, we motivate our method to make the classification scheme feasible on large-scale bisample data. To this end, its robustness to bisample data and scalability to large-scale classes should be enhanced. First, we find the classification scheme convergent on bisample data only if it is well initialized. So that we propose a CVC training strategy to initialize the model and construct the prototypes for the classification scheme. Second, we propose a prototype selection strategy to scale up the classification scheme to any number of classes. With the improvements, we achieve superior performance to existing methods.

\subsection{Bisample Learning}
\label{sec:strategy and exempalr construction}
It has been observed that when training data is insufficient, transferring knowledge from related tasks is better than directly training on the target domain~\cite{koch2015siamese}. Inspired by this, we regard the well-posed wild data as the source domain and the IvS data as the target domain. A classification-verification-classification (CVC) training strategy is proposed to transfer the knowledge from wild scenarios to IvS scenarios and boost the performance by large-scale classification. As shown in Fig.~\ref{fig-overview}, the CVC involves three stages:
\begin{enumerate}
\item \textbf{Pre-learning (Classification)}: We first train the deep model on a wild dataset to get a good initialization for general face recognition. With a limited number of classes (less than $100,000$), we can adopt a classification loss like softmax~\cite{Sun-CVPR-2014} and A-softmax~\cite{liu2017sphereface} to perform one-vs-all optimization. The trained model performs well in wild scenarios but terribly in IvS scenarios due to the large bias~\cite{zhou2015naive}. Nevertheless, the model has learned basic knowledge about human faces and will not be puzzled by IvS data.
\item \textbf{Transfer Learning (Verification)}: Since the verification scheme only concerns a small number of classes and just needs two samples per class to optimize intra-class distances in each iteration. We believe verification is robust to large-scale bisample data. In this stage, we adopt the verification scheme to transfer the face knowledge from wild scenarios to IvS scenarios.
    Specifically, we remove the classification layer and finetune the model on the IvS dataset with a verification loss like contrastive~\cite{sun2014deep} or triplet~\cite{schroff2015facenet}. Benefiting from the initialization from the previous stage and the robustness to bisample data of the verification scheme, we can successfully optimize the loss function and provide a good initialization for the final large-scale classification.
\item \textbf{Fine-grained Learning (Classification)}: We construct a classification layer on the top of the network and conduct classification with $2$ million classes on the IvS dataset. A novel dominant prototype softmax is adopted to select a small number of dominant classes to participate into the classification in each iteration. The new softmax can effectively and efficiently perform large-scale classification and further boost the performance, finally achieves satisfactory recognition accuracy in IvS scenarios.
\end{enumerate}

The key in CVC is that the knowledge transferring should be smooth. We find after the first stage, the large-scale classification has been able to converge. However, the loss descends slowly and the optimization gets stuck into a bad local optima. Considering that the verification scheme has good robustness to data distribution, we bridge the two classification stages with a verification stage, which gives a better initialization for large-scale classification and finally achieves much better performance. Although classification followed by verification~\cite{parkhi2015deep} and the joint identification-verification~\cite{sun2014deep} have been applied in training web-face models, the two schemes are applied on the same dataset. While the first two stages of CVC are applied on different datasets with different scenarios, which acts as a knowledge-transferring role.

To perform classification in the final stage of CVC, we must construct the absent classification layer, which contains the prototype for each class. Considering prototypes serve as the class proxies, to which the deep features will be optimized, we construct the prototype of a class by the features belonging to it. Specifically, we try two kinds of prototypes: ID-prototype and avg-prototype. Suppose $\mathbf{x}^{id}_{i}$ and $\mathbf{x}^{spot}_{i}$ are the deep features of the ID and spot photos of the $i$th identity, we set the ID-prototype $\mathbf{w}^{id}_{i}=\mathbf{x}^{id}_{i}$ and the avg-prototype $\mathbf{w}^{avg}_{i}=(\mathbf{x}^{id}_{i} + \mathbf{x}^{spot}_{i})/2$. Intuitively, the ID-prototype enforces the spot feature to approach the more reliable ID feature and the avg-prototype makes the two features approach their centroid. Our experiments show that which kind of prototype is better depends on the loss function.

In the next section, we will introduce how to perform large-scale classification in the final stage of CVC.

\subsection{Large-scale Classification}
\label{sec:exempalr selection}
\subsubsection{Random Prototype Softmax}
\label{subsec: random exempalr selection}
With the well initialized network and prototypes, the only problem remained is to scale up the classification scheme to massive classes. If we directly perform classification on $2$ million classes, the massive prototypes will take $1/3$ GPU memory ($4$GB of the $12$GB) and dramatically increase the training time due to their numerous parameters.

\begin{figure*}[!htb]
  \centering{}
  \includegraphics[width=0.95\textwidth]{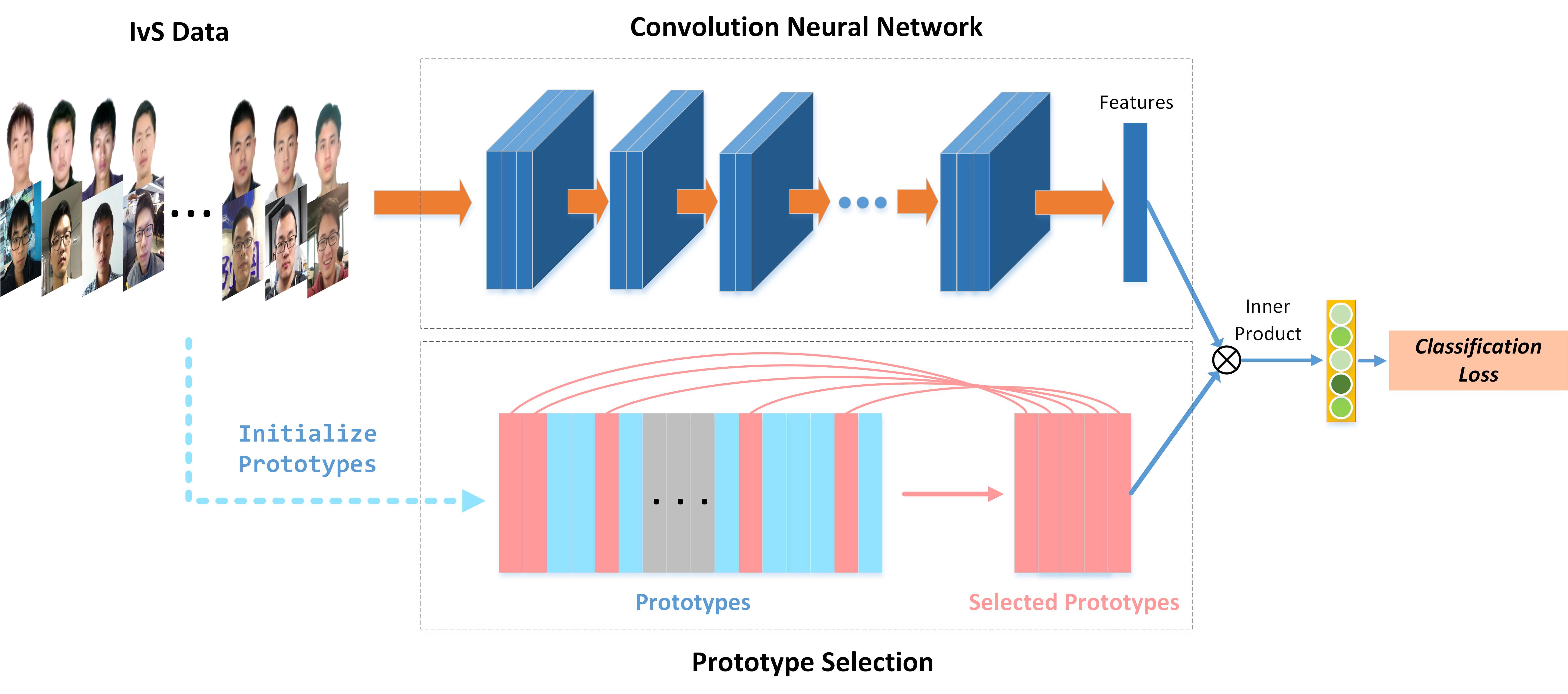}{}
  \caption{Overview of large-scale classification.}
  \label{fig-prototype-selection}
\end{figure*}

We aim to improve scalability by reducing the cost of large-scale classification. As shown in Fig.~\ref{fig-prototype-selection}, we select a fraction of prototypes to participate in the classification in each iteration. In the pair-matching formulation of softmax (Equ.~\ref{equ-dummy-softmax}), given one mini-batch $\mathbf{X}=[\mathbf{x}_{1},\ldots,\mathbf{x}_{M}]$ where samples have different labels, all the prototypes $\mathbf{W}=[\mathbf{w}_{1},\ldots,\mathbf{w}_{N}]$ can be divided into $M$ positive prototypes $\mathbf{W}_{pos}$ and the rest negative prototypes $\mathbf{W}_{neg}$. Each prototype in $\mathbf{W}_{pos}$ has a mate in $\mathbf{X}$ to make up a positive pair, while the prototypes in $\mathbf{W}_{neg}$ do not share class with any of $\mathbf{X}$ and only make up negative pairs. Given that $M \ll (N-M)$, it is unnecessary to put the whole $\mathbf{W}_{neg}$ into GPU memory since negative pairs are redundant. Based on this, we propose a naive solution called \textbf{Random Prototype Softmax (RP-softmax)}. The RP-softmax stores the full prototype matrix $\mathbf{W}$ in the memory. In each iteration, it first constructs a temporary prototype matrix $\mathbf{W}_{iter}=[\mathbf{W}_{pos},\widehat{\mathbf{W}}_{neg}]$, where $\widehat{\mathbf{W}}_{neg}$ has $N_{iter}-M$ randomly selected prototypes from $\mathbf{W}_{neg}$ and $N_{iter}$ is the number of selected prototypes. Then $\mathbf{W}_{iter}$ is copied into GPU for training and updated to $\mathbf{W}_{iter}^{+}$. Finally, $\mathbf{W}_{iter}^{+}$ and $\mathbf{W}$ are synchronized by replacing the selected prototypes with the updated ones. Overall, the prototype selection and updating procedure is listed in Algorithm~\ref{algo:exemplar selection random}.

\begin{algorithm}[!htb]
\caption{Random Prototype Softmax \label{algo:exemplar selection random}}
\SetKwInOut{Input}{Input}\SetKwInOut{Output}{Output}
\SetKwInput{Initialization}{Initialization}
\SetKwFunction{Median}{Median}
\Input{ Prototype matrix: $\mathbf{W}=[\mathbf{w}_{1},\ldots,\mathbf{w}_{N}]$ \\
        ~~Feature matrix: $\mathbf{X}=[\mathbf{x}_{1},\ldots,\mathbf{x}_{M}]$ \\
        ~~Number of selected classes $N_{iter}$}
\Output{ Updated prototype matrix }
Initialize selected class set $\mathcal{L}=\emptyset$ \\
\For{each feature $\mathbf{x}_{j}$ in $\mathbf{X}$}{
   Get the label $y^{(j)}$ of $\mathbf{x}_{j}$ \\
   $\mathcal{L}$.insert($y^{(j)}$)  \\
}
\While{$\mathcal{L}.size < N_{iter}$}{
  randomly select a label $y^{(r)}$ \\
  $\mathcal{L}$.insert($y^{(r)}$) \\
}
\For{$i = 1 \ldots N_{iter}$}{
  $\mathbf{W}_{iter}[i,:] = \mathbf{W}[\mathcal{L}_{i},:]$, $\mathcal{L}_{i}$ is the member of $\mathcal{L}$\\
}
Training with $\mathbf{W}_{iter}$ and $\mathbf{X}$, getting updated $\mathbf{W}_{iter}^{+}$ \\
\For{$i = 1 \ldots N_{iter}$}{
  $\mathbf{W}[\mathcal{L}_{i},:]$ = $\mathbf{W}_{iter}^{+}[i,:]$\\
}
\end{algorithm}

The hyper parameter $N_{iter}$ plays a key role in RP-softmax. Larger $N_{iter}$ brings more negative pairs and provides richer inter-variation information. However, increasing $N_{iter}$ is not cost free. Besides the time-consuming large matrix multiplication, the softmax layer has to get blocked until $\mathbf{W}_{iter}$ is copied into GPU. Sometimes the waiting time exceeds the forward propagation time. Moreover, increasing $N_{iter}$ squeezes the batch size and degrades the data-driven layers like batch-normalization. As a result, $N_{iter}$ is set empirically to balance the performance and the training time. In our experiments, with $N_{iter} = 100,000$ the RP-softmax significantly improves the performance in IvS scenarios.

\subsubsection{Dominant Prototype Softmax}
\label{subsec: dominant queue selection}

Although RP-softmax makes it possible to perform large-scale classification, it is still inefficient due to its blind prototype selection. In this section, we show that the quality not the quantity really matters in prototype selection. We begin with the demonstration that in each iteration, only a small fraction of negative prototypes generate strong gradients.

\begin{figure}[!htb]
  \centering{}
  \includegraphics[width=0.50\textwidth]{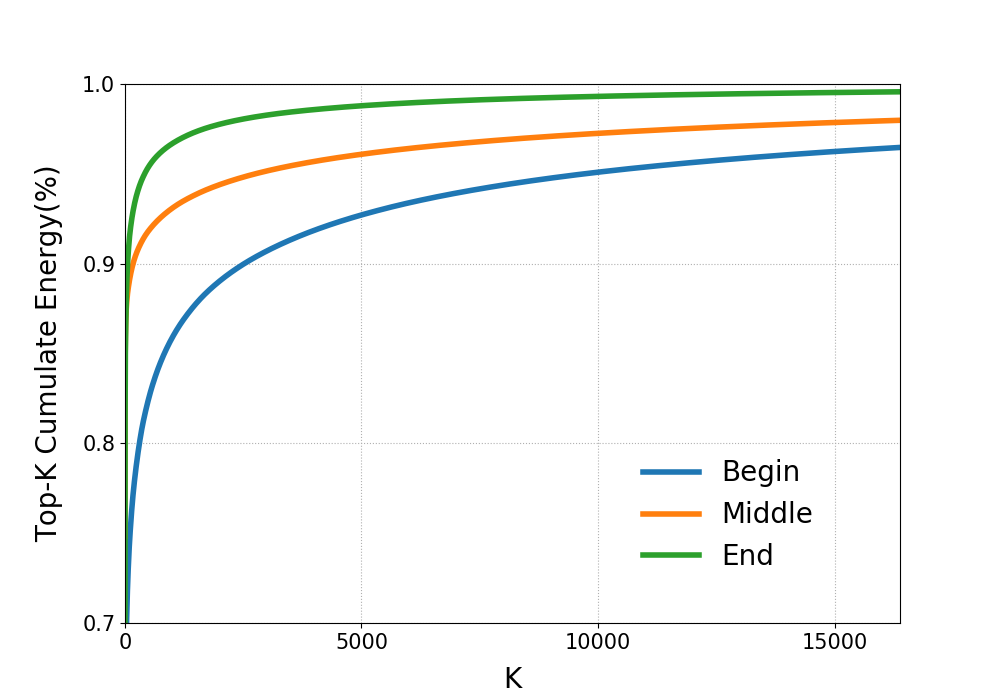}{}
  \caption{The top-$K$ cumulate energy of negative prototypes ($\emph{CE}_{K}$) for a mini-batch, in the beginning, middle ($100,000$ iterations) and end ($200,000$ iterations) of the training process. The batch size is $50$ and the number of classes is $2,578,178$. The curves come from averaging $\emph{CE}_{K}$ of $5,000$ mini-batches.}
  \label{fig-cek}
\end{figure}

In Equ.~\ref{equ-softmax-derivative}, a negative prototype $\mathbf{w}_{i}$ contributes to the back-propagated gradient by $p_{ij}\mathbf{w}_{i}$, whose norm is $p_{ij}\|\mathbf{w}_{i}\|$. Usually, we restrict $\|\mathbf{w}_{i}\|$ to one~\cite{liu2017sphereface} and the norm will be $p_{ij}$, which can measure the impact of $\mathbf{w}_{i}$ to the training process. In this paper, with a mini-batch $\mathbf{X}=[\mathbf{x}_{1},\ldots,\mathbf{x}_{M}]$, we define the energy of a negative prototype as:
\begin{equation}\label{equ-neg-energy}
\emph{E}_{neg}(\mathbf{w}_{i}) = \sum_{j=1}^{M}p_{ij},
\end{equation}
where $p_{ij}$ is the probability that $\mathbf{x}_{j}$ belongs to class $i$. Note that none of $\mathbf{X}$ has the label $i$ since $\mathbf{w}_{i}$ is a negative prototype. To analyze whether the energy is concentrated on a small fraction of prototypes, we further define the top-$K$ cumulative energy as:
\begin{equation}\label{equ-topk-ner}
\emph{CE}_{K} = \frac{\sum_{\mathbf{w}_{i} \in \mathcal{T}_{K}} \emph{E}_{neg}(\mathbf{w}_{i})}{\sum_{\mathbf{w}_{i} \in \mathcal{W}_{neg}} \emph{E}_{neg}(\mathbf{w}_{i})},
\end{equation}
where $\mathcal{W}_{neg}$ is the set of negative prototypes and $\mathcal{T}_{K}$ is the set of $K$ negative prototypes with the largest energy. A large $\emph{CE}_{K}$ with small $K$ denotes that the energy of negative prototypes are highly concentrated. We plot the $\emph{CE}_{K}$ along the training process in Fig.~\ref{fig-cek}. It can be seen that in the beginning the top-$5000$ possesses $92.71\%$ of energy. As the training proceeds, the energy becomes more and more concentrated. In the middle and end of the training process, the energy of top-$5000$ is increased to $96.09\%$ and $98.79\%$. These results indicate that only a small fraction of prototypes can produce large gradients to affect training. We call these negative prototypes with large energy as \textbf{dominant prototypes}.

In real implementation, given a batch of features, how can we know the most dominant prototypes before we compute the probabilities in softmax? In this paper, we assume that if two identities have similar ID features, their prototypes and features are likely to make hard negative pairs.
Based on this, we propose the \textbf{Dominant Prototype Softmax (DP-softmax)}. The basic idea is selecting prototypes from a set of dominant queues and updating the queues by the softmax predications. The procedure is detailed as follows:

\textbf{Queue Initialization}: For each class $i$, we define the K-\textbf{N}earest \textbf{C}lasses $\mathcal{NC}_{K}^{(i)}$ as the top-$K$ classes having the nearest ID features with $i$. Before training, we build an approximate nearest neighbor (ANN) graph by ID features and get the $\mathcal{NC}_{K}^{(i)}$ for each class. Then we construct a dominant queue $\mathcal{Q}_{i}$ and a candidate set $\mathcal{C}_{i}$ for each class. The $\mathcal{Q}_{i}$ is initialized by $\mathcal{NC}_{100}^{(i)}$ and its members are sorted by the distances of ID features to $i$. The $\mathcal{C}_{i}$ is set to $\mathcal{NC}_{300}^{(i)}$. Note that $\mathcal{Q}_{i} \subseteq \mathcal{C}_{i}$.

\textbf{Prototype Selection}: After training begins, in each iteration we need to select prototypes for the mini-batch $\mathbf{X}=[\mathbf{x}_{1}, \ldots, \mathbf{x}_{M}]$. First we select their positive prototypes $\mathbf{W}_{pos}=[\mathbf{w}_{y^{(1)}},\ldots,\mathbf{w}_{y^{(M)}}]$ where $y^{(j)}$ is the label of $\mathbf{x}_{j}$. Second, for each feature $\mathbf{x}_{j}$ we select the prototypes of the classes in its dominant queue that $\mathbf{W}^{\text{qu}}_{j}=[\mathbf{w}_{q} | q\in \mathcal{Q}_{y^{(j)}}]$ and the full negative prototypes are $\widehat{\mathbf{W}}_{neg}=[\mathbf{W}^{\text{qu}}_{j},\ldots,\mathbf{W}^{\text{qu}}_{M}]$. Thirdly, we remove the repeated prototypes and randomly select negative prototypes into $\widehat{\mathbf{W}}_{neg}$ until a preset number is reached. Finally $\mathbf{W}_{pos}$ and $\widehat{\mathbf{W}}_{neg}$ constitute the temporary prototype matrix $\mathbf{W}_{iter}$ in this iteration and are copied into GPU for training. Algorithm~\ref{algo:dominant-prototype-softmax} summarizes the DP-softmax.

\begin{algorithm}[!htb]
\caption{Dominant Prototype Softmax \label{algo:dominant-prototype-softmax}}
\SetKwInOut{Input}{Input}\SetKwInOut{Output}{Output}
\SetKwInput{Initialization}{Initialization}
\SetKwFunction{Median}{Median}
\Input{ Prototype matrix: $\mathbf{W}=[\mathbf{w}_{1},\ldots,\mathbf{w}_{N}]$ \\
        ~~Feature matrix: $\mathbf{X}=[\mathbf{x}_{1},\ldots,\mathbf{x}_{M}]$ \\
        ~~Number of selected classes $N_{iter}$ \\
        ~~Dominant Queues: $\mathcal{Q}_{i}, i = 1,\ldots,N$}
\Output{ Updated prototype matrix}
Initialize selected class set $\mathcal{L}=\emptyset$ \\
\For{each feature $\mathbf{x}_{j}$ in $\mathbf{X}$}{
   Get the label $y^{(j)}$ of $\mathbf{x}_{j}$ \\
   $\mathcal{L}$.insert($y^{(j)}$)  \\
   \For{each class $k$ in $\mathcal{Q}_{y^{(j)}}$}{
        $\mathcal{L}$.insert($k$)
   }
}
\While{$\mathcal{L}.size < N_{iter}$}{
  randomly select a label $y^{(r)}$ \\
  $\mathcal{L}$.insert($y^{(r)}$) \\
}
\For{$i = 1 \ldots N_{iter}$}{
  $\mathbf{W}_{iter}[i,:] = \mathbf{W}[\mathcal{L}_{i},:]$, $\mathcal{L}_{i}$ is the member of $\mathcal{L}$\\
}
Training with $\mathbf{W}_{iter}$ and $\mathbf{X}$, getting updated $\mathbf{W}_{iter}^{+}$ \\
\For{$i = 1 \ldots N_{iter}$}{
  $\mathbf{W}[\mathcal{L}_{i},:]$ = $\mathbf{W}_{iter}^{+}[i,:]$\\
}
\end{algorithm}

\begin{figure}[!htb]
  \centering{}
  \subfigure[Mislabelling]{
  \includegraphics[width=0.23\textwidth]{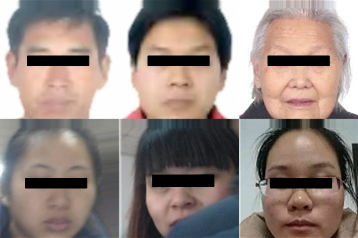}}
  \subfigure[Low Quality]{
  \includegraphics[width=0.23\textwidth]{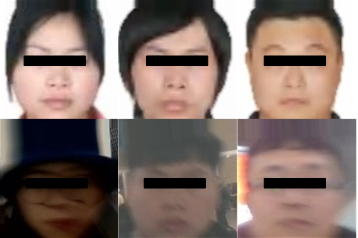}}
  \caption{The refused mis-predicted class. When a mis-predicted class is refused to enter the dominant queue, there are always something wrong in its spot photo, including (a) mislabelling and (b) low quality.}
  \label{fig-queue-update-mis}
\end{figure}

\textbf{Queue Updating}: After training in each iteration, we can update the dominant queues by the predictions of softmax. For a feature $\mathbf{x}_{j}$, its highest
activated class $h$ provides valuable information: First if $h=y^{(j)}$ then it is a successful prediction and there is nothing to update. Second if $h \neq y^{(j)}$ but $h\in\mathcal{Q}_{y^{(j)}}$, then this is a mis-prediction but the wrong-matched class is still in the dominant queue. Hence we need not to update $\mathcal{Q}_{y^{(j)}}$. Thirdly, if $h \neq y(j)$ and $h\not\in\mathcal{Q}_{y^{(j)}}$ but $h\in\mathcal{C}_{y^{(j)}}$, it means the class neighborhood has changed as the training proceeds. Therefore, we push $h$ into $\mathcal{Q}_{y^{(j)}}$ and pop the class that is the most dissimilar to $y^{(j)}$. Finally if $h \neq y^{(j)}$ and $h$ is not in $\mathcal{Q}_{y^{(j)}}$ or $\mathcal{C}_{y^{(j)}}$, it means $h$ and $y^{(j)}$ have dissimilar ID features in the beginning but become close at this time. This case is mostly caused by the mislabelled or low-quality spot photo of $h$ which misdirects its prototype, as shown in Fig.~\ref{fig-queue-update-mis}. Therefore, we do not update $\mathcal{Q}_{y^{(j)}}$ since $h$ is a noisy label.

The whole prototype selecting and queue updating operations can be done in real time. Compared with the RP-softmax, the DP-softmax significantly improves the quality and reduce the quantity of prototypes, leading to faster training and better performance.

Since the prototypes are saved in memory, which can easily hold tens of millions of prototypes, the dominant prototype selection scales up the classification scheme to any number of classes. Besides, when new training data come, the prototype matrix $\mathbf{W}$ can be extended by the ID features of the new identities. Then the network can be finetuned on the whole training data.

\section{Experiments}\label{sec-experiments}
In this section, the proposed large-scale bisample learning (LBL) is systematically evaluated. We first analyze the CVC training strategy. Then we explore how different prototype selection methods affect the final performance. Finally we conduct comparison experiments on three datasets including CASIA-IvS-Test, Public-IvS and Megaface-bisample.

\subsection{Datasets}
\textbf{Ms-Celeb-1M}: The Ms-Celeb-1M~\cite{guo2016ms} is one of the largest wild dataset containing $98,685$ celebrities and $10$ million images. The list of~\cite{wu2015light} is adopted to clean the noisy labels, resulting in $79,077$ identities and $5$ million images.

\begin{figure}[!htb]
  \centering{}
  \includegraphics[width=0.48\textwidth]{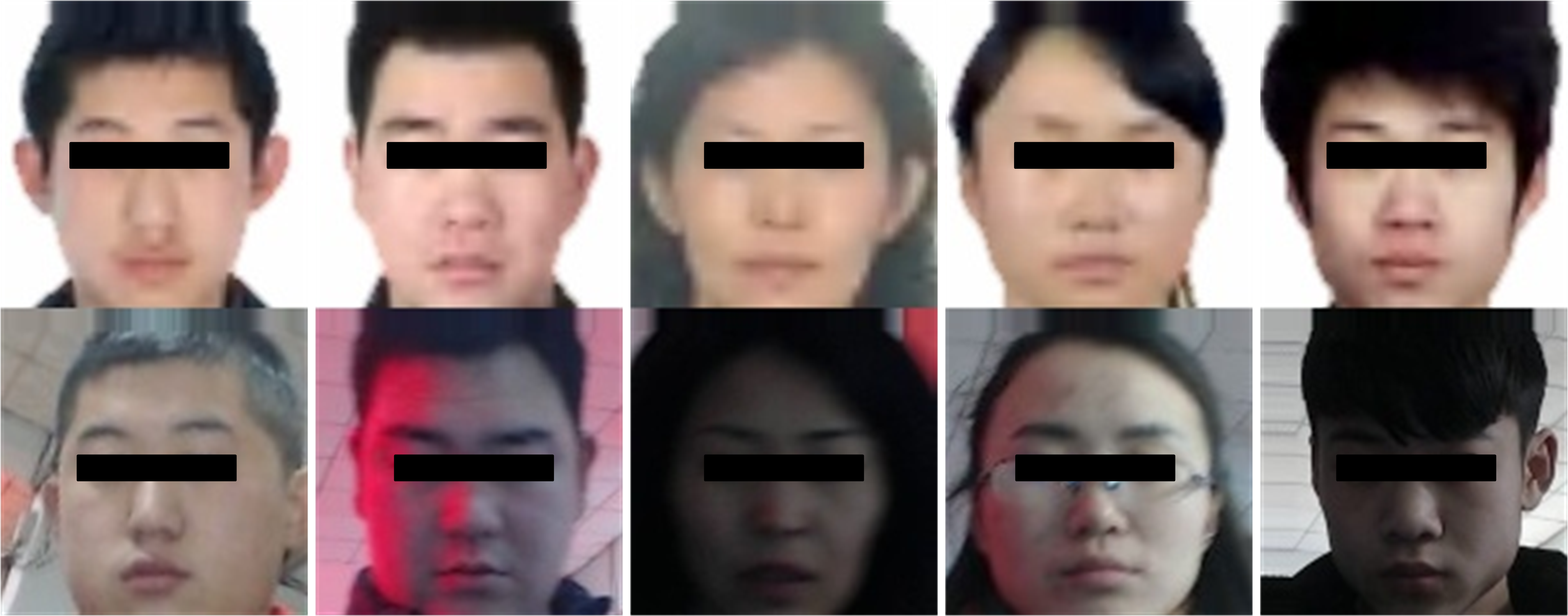}{}
  \caption{Example images in CASIA-IvS.}
  \label{fig-demo-casia-ivs}
\end{figure}

\textbf{CASIA-IvS}: The CASIA-IvS dataset is collected for IvS face recognition. The training set CASIA-IvS-Train contains $2,578,178$ identities, each having two images. One image is the ID photo from the ID card, which is taken with uniform background, in frontal pose, normal illumination and neutral expression. The other is the spot photo taken by on-site devices, with variations in pose, expression, illumination, occlusion and resolution, as shown in Fig.~\ref{fig-demo-casia-ivs}. The test set CASIA-IvS-Test contains $4,000$ identities and $8,000$ images, which are checked manually to clean the noisy labels and ensure there is no identity overlap between training and test sets. During testing, all the ID photos and spot photos are paired, generating $4,000$ positive pairs and nearly $16$ million negative pairs.

\begin{figure}[!htb]
  \centering{}
  \includegraphics[width=0.48\textwidth]{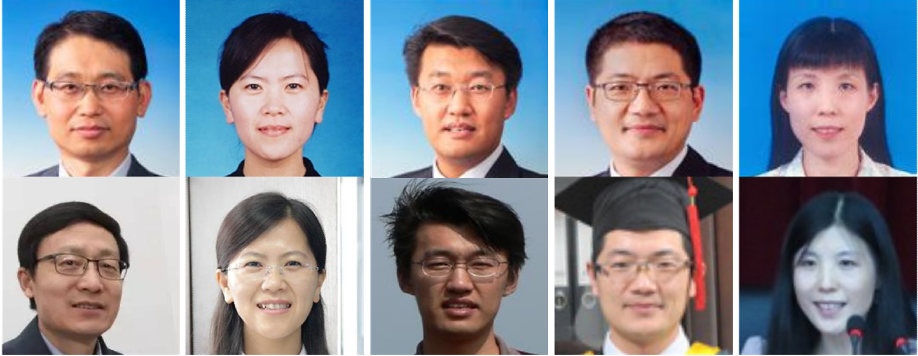}{}
  \caption{Example images in Public-IvS.}
  \label{fig-demo-public-ivs}
\end{figure}

\textbf{Public-IvS}: An IvS test dataset is released for open evaluation. We found some public characters, such as politicians, teachers and researchers, had their ID photos on BaiduBaike~\cite{url-baidubaike} and official pages. We recorded their names and collected their spot photos on the web. Afterwards, we cleaned the dataset manually and removed the profile-view images. The final Public-IvS dataset has $1,262$ identities and $5,507$ images, each identity having one ID photo and $1$ to $10$ spot photos. There are $4,871$ positive pairs and nearly $6$ million negative pairs. Fig.~\ref{fig-demo-public-ivs} shows some images in Public-IvS. Although Public-IvS is not a strictly IvS dataset since the spot photos are collected from the web, experiments on Public-IvS have consistent results with the real-world CASIA-IvS-Test.

\subsection{Experimental Settings}\label{sec-implemental-details}

\textbf{Preprocessing}~~ We detect faces by the FaceBox~\cite{zhang2017faceboxes} detector and localize $5$ landmarks (two eyes, nose tip and two mouth corners) by a simple $6$-layer CNN~\cite{feng2017wing}. All the faces are normalized by similarity transformation and cropped to $120 \times 120$ RGB images.

\textbf{CNN Architecture}~~ For the sake of fairness, all the CNN models in the experiments follow the same ResNet64 architecture~\cite{liu2017sphereface}. It has four residual blocks and gets a $512$-dimensional feature vector by average pooling. The learning rate begins with $0.001$ and is divided by $10$ when the loss does not decrease. All the networks are trained on $4$ TITANX GPUs parallelly and the batch size is set to occupy all the GPU memory. Specifically, the batch size is $66$ in the verification scheme and about $50$ in the classification scheme.

\textbf{Training Setup}~~ There are three stages in the CVC training strategy: pre-learning by classification on wild data, transfer learning by verification on IvS data and fine-grained learning by large-scale classification on IvS data. In the first stage, we train model from scratch by the A-Softmax loss~\cite{liu2017sphereface} on the Ms-Celeb-1M. In the second stage, we finetune the model on CASIA-IvS-Train with the triplet loss~\cite{schroff2015facenet}. The triplet loss is modified by N-pairs batch construction~\cite{sohn2016improved}, online hard-negative mining~\cite{schroff2015facenet} and anchor swapping~\cite{balntas2016learning}. In the third stage, we adopt the proposed DP-softmax to finetune the model on CASIA-IvS-Train. If not specified, there are two samples for each class in a mini-batch; the classification layer in the third stage is initialized by the ID-prototypes; softmax provides the probabilities and A-softmax provides the gradients. In DP-softmax the sizes of dominant queues and candidate sets are $100$ and $300$, respectively.

\textbf{Evaluation Setup}~~ For each image, we extract features from both the original image and the flipped one and concatenate them as the final representation. The score is measured by the cosine distance of two features. We evaluate all the networks with ROC curves. The verification rate (VR) at low false acceptance rate (FAR) is preferred since in real application false acceptance gives higher risks than false rejection.

\begin{table*}[!tb]
  \tabcolsep 10pt
  \setlength{\tabcolsep}{5pt}
  \footnotesize
  \begin{center}
    \begin{tabular}{@{}c|c|c|c||c|c|c@{}}
      \hline
      {\multirow{3}{*}{Method}} & \multicolumn{3}{c||}{Procedure}  & \multicolumn{3}{c}{Performance}\\
      \cline{2-7}
      {\multirow{3}{*}{}}&  Classification & Verification & Classification & {\multirow{2}{*}{VR@FAR=$10^{-3}$}} & {\multirow{2}{*}{VR@FAR=$10^{-4}$}} & {\multirow{2}{*}{VR@FAR=$10^{-5}$}} \\
      {\multirow{3}{*}{}} &(A-Soft on MS) & (Triplet on IvS) & (DP-Soft on IvS) & {\multirow{2}{*}{}} & {\multirow{2}{*}{}} & {\multirow{2}{*}{}} \\
      \hline
      C\#\# & $\checkmark$ &  &  & 85.31 & 69.11 & 51.90    \\
      \hline
      CV\# & $\checkmark$ & $\checkmark$ &  & 96.41 & 91.39 & 83.23  \\
      \hline
      CVC & $\checkmark$ & $\checkmark$ & $\checkmark$ & 97.70 & 96.17 & 91.92 \\
      \hline
      \hline
      \#\#C & &  & $\checkmark$ & \multicolumn{3}{c}{ not converge } \\
      \hline
      \#V\# & & $\checkmark$ &  & 79.39 & 58.90 & 38.33 \\
      \hline
      C\#C &$\checkmark$ &  & $\checkmark$ & 94.36 & 85.35 & 72.35  \\
      \hline
      C(VC) &$\checkmark$ & \checkmark & $\checkmark$ &  96.43 & 91.75  & 82.80   \\
      \hline
    \end{tabular}
  \end{center}
  \caption{The intermediate results after each stage in the CVC training strategy. The performance is evaluated by the verification rate, VR(\%), on CASIA-IvS-Test. In each stage, we indicate the loss function and the training data, where A-Soft refers to A-softmax, Triplet refers to triplet loss, DP-Soft refers to DP-softmax, MS refers to Ms-Celeb-1M and IvS refers to CASIA-IvS-Train. The ``\#'' in method names indicates the corresponding stage is skipped.}
  \label{table-training-strategy}
\end{table*}

\subsection{Bisample Training}\label{sec:exp-examplar-softmax}

\subsubsection{Classification-Verification-Classification (CVC)}\label{sec:experiment-cvc}
To illustrate the effectiveness of CVC, we show the intermediate results in Table~\ref{table-training-strategy}. After the first stage, the C\#\# is a well trained model in wild scenarios, with $99.53\%$ on LFW~\cite{Huang-2007-LFW} and $90.38\%$ at FAR$=10^{-6}$ on Megaface challenge~\cite{nech2017level}. However the state-of-the-art face model cannot work well on CASIA-IvS-Test, indicating the large bias between the two scenarios. Second, after being finetuned on CASIA-IvS-Train with the triplet loss, the CV\# achieves much better performance, indicating the knowledge is successfully transferred from wild scenarios to IvS scenarios. Finally, the large-scale classification on CASIA-IvS-Train further improves the performance and reaches $91.92\%$ at FAR=$10^{-5}$.

To further analyze the impact of each stage, we perform an ablation study by removing some stages. First, in \#\#C we directly perform large-scale classification on IvS data without any initialization and find the loss does not decrease after $200,000$ iterations. Second, we try to train model from scratch by the triplet loss on IvS data. Since the learning task is challenging without any initialization, we begin without hard-negative mining and slightly increase the ratio of hard negatives. The training converges but the model \#V\# has a bad result. Thirdly, we pre-train the model on wild data and directly finetune it on IvS data by large-scale classification. The training successfully converges but the resultant C\#C is worse than the complete CVC. Finally, after pre-training on wild data, we perform joint verification and large-scale classification on IvS data, yielding the C(VC) model, which is also inferior than the complete CVC. From the results we can conclude that: (1) Comparing \#\#C, C\#C and CVC, a good initialization is crucial for the large-scale classification on bisample data. (2) Comparing C\#C, CV\# and CVC, the verification scheme has higher scalability than the classification scheme when dealing with large-scale bisample data, but it cannot get satisfactory performance independently. (3) Comparing C\#C, C(VC) and CVC, the smoothness is important in knowledge transferring and it is better to bridge the two classification stages with a verification stage.

There are some interesting phenomena we have observed in CVC learning. First, we find that the wild performance in the first stage does not affect the final IvS performance much. We begin with two pre-trained models with different wild performance ($98.0\%$ on LFW with triplet loss and $99.53\%$ with A-softmax) and find their final IvS performances differ slightly ($91.23\%$ vs. $91.92\%$ at FAR$=10^{-5}$ on IvS). Second, we find the model cannot keep its high wild performance after being finetuned on IvS data. We evaluate models on both CASIA-IvS-Test and LFW~\cite{liao2014benchmark}, shown in Table~\ref{table-training-strategy-onweb}. After each stage of CVC, the IvS performance is improved at the cost of degenerated wild performance. We further train our model on the joint data from both scenarios and find the wild performance is greatly improved with slight drop in IvS. This joint training is a good strategy when both scenarios are concerned.

\begin{table}[!htb]
  \footnotesize \hspace{-2mm}
    \resizebox{\columnwidth}{!}{%
    \begin{tabular}{|c||c|c|}
      \hline
      {\multirow{2}{*}{Methods}} & CASIA-IvS-Test & LFW-BLUFR \\
       & (VR@FAR=$10^{-5}$) &(VR@FAR=$10^{-5}$) \\
      \hline
      C\#\#       & 51.90 & 94.23  \\
      \hline
      CV\#       & 83.23 & 86.38  \\
      \hline
      CVC       & 91.92 & 80.71  \\
      \hline
      CVC$^{+}$ & 89.96 & 90.81  \\
      \hline
    \end{tabular}
    }
  \caption{The performances in wild scenarios (LFW-BLUFER protocol~\cite{liao2014benchmark}) and IvS scenarios (CASIA-IvS-Test) after each stage CVC. The CVC$^{+}$ means the final large-scale classification stage is performed on the joint data from both Ms-Celeb-1M and CASIA-IvS-Train.}
  \label{table-training-strategy-onweb}
\end{table}


\subsubsection{Prototype Construction}\label{sec:experiment-construction-loss}
As introduced in Section~\ref{sec:strategy and exempalr construction}, there are two ways to construct the prototypes in large scale classification: The ID-prototype is the feature of the ID photo and the avg-prototype is the average vector of all the features in this class. The way to construct prototypes depends on the loss function involved. We select the most representative softmax~\cite{Sun-CVPR-2014} and the state-of-the-art A-softmax~\cite{liu2017sphereface} in this experiment. Table~\ref{tab-examplar-construction} shows the performances with different losses and prototypes.
\begin{table}[!htb]
  \resizebox{\linewidth}{!}{%
    \begin{tabular}{|c||c|c|c|@{}}
     \hline
      Method & FAR=$10^{-3}$ & FAR=$10^{-4}$ & FAR=$10^{-5}$  \\
      \hline
      softmax (avg) & 96.94 & 93.55 & 87.13   \\
      softmax (ID)  & 97.31 & 94.91 & 89.55   \\
      \hline
      A-softmax (avg) & 97.35 & 95.40 & 90.30  \\
      A-softmax (ID)  & 97.43 & 95.40 & 90.34  \\
      \hline
    \end{tabular}
    }
  \caption{The comparison of different prototype construction methods with different loss functions on CASIA-IvS-Test, evaluated by VR(\%) at different FAR. The prototypes are randomly selected.}
  \label{tab-examplar-construction}
\end{table}

When softmax is adopted, the model initialized by avg-prototypes almost converges in the beginning and the loss only produces small gradients. If we replace avg-prototypes with ID-prototypes, the softmax loss will have a larger initial loss and end up with better results.
When A-softmax is adopted, the angular margin keeps the initial loss large enough and the two prototypes end up with close performances. In our experiments, we prefer ID-prototypes and only adopt avg-prototypes when there is no ID photo like the mimic experiments in Sec.~\ref{sec:exp-mimic}.

\subsection{Large-scale Classification}\label{sec-exp-examplar-selection}
In large-scale classification, we need to select a fraction of prototypes each time. In Sec.~\ref{sec:exempalr selection} we introduce two methods for prototype selection: one is to select prototypes randomly and the other is to select the dominant prototypes.

\subsubsection{Random Prototype Softmax}\label{sec:exp-random-selection}
In random prototype softmax (RP-softmax), we can increase the involved classes at a small cost of batch size due to the tiny memory cost of a single prototype. We evaluate the RP-softmax with $20$k, $50$k and $100$k prototypes respectively in Table~\ref{tab-perf-random-prototype} and find more prototypes always come with better performance.

\begin{table}[!htb]
  \footnotesize \hspace{-2mm}
    \resizebox{\linewidth}{!}{%
    \begin{tabular}{|c||c|c|c|}
      \hline
      Method & FAR=$10^{-3}$ & FAR=$10^{-4}$ & FAR=$10^{-5}$ \\
      \hline
      RPS(20k) & 97.19 & 94.49 & 87.84  \\
      RPS(50k)  & 97.19 & 94.71 & 88.57  \\
      RPS(100k) & 97.43 & 95.40 & 90.34  \\
      \hline
    \end{tabular}
    }
  \caption{The performance of RP-softmax (RPS) on CASIA-IvS-Test, evaluated by VR(\%) at different FAR. The values in the brackets are the numbers of prototypes.}
  \label{tab-perf-random-prototype}
\end{table}

\begin{figure}
  \centering{}
  \includegraphics[width=0.5\textwidth]{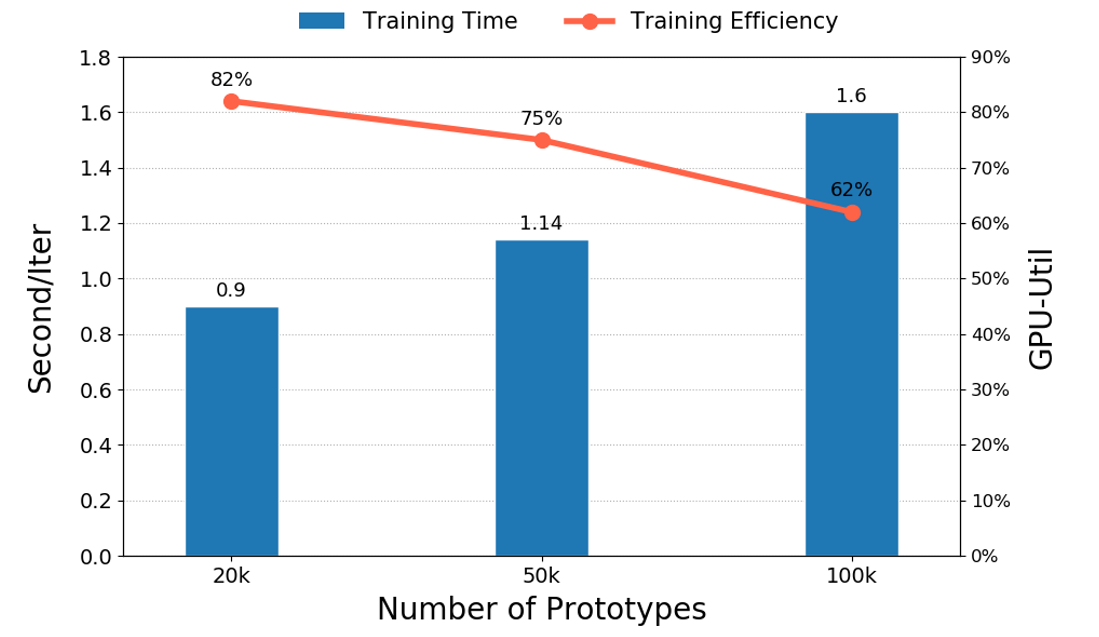}{}
  \caption{The total training time (forward and backward propagation) of one mini-batch and the GPU-util percent with different prototype numbers. Low GPU-util percent means the GPU is blocked to wait for prototype copying.}
  \label{fig:time-random-prototype}
\end{figure}

However, increasing the number of prototypes is not cost free. More prototypes increase the overhead of computing softmax and copying prototypes in GPUs. In Fig.~\ref{fig:time-random-prototype}, we show the time costs and GPU-util percent with different prototype numbers. When prototypes increase from $20$k to $100$k, the training time increases by $78\%$ and the GPU-util percent drops from $82\%$ to $62\%$. We further try $300$k prototypes and find the GPU-util percent drops to $48\%$, which means most time is spent on waiting for prototype copying.

\subsubsection{Dominant Prototype Softmax}\label{sec:exp-smart-selection}
\begin{figure}
  \centering{}
  \includegraphics[width=0.5\textwidth]{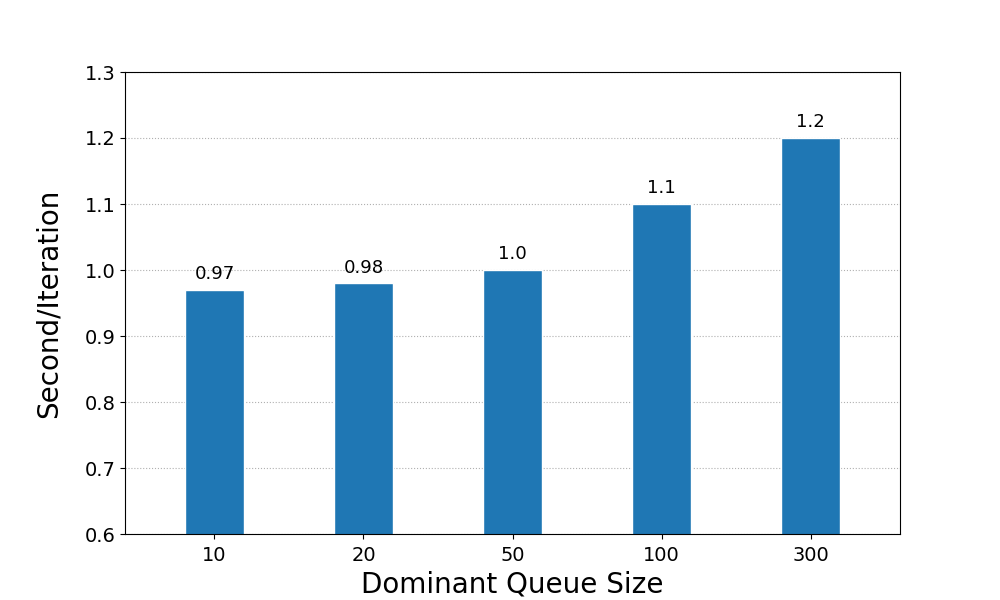}{}
  \caption{The total training time (forward and backward propagation) of one mini-batch with different dominant queue size.}
  \label{fig:time-dominant-prototype}
\end{figure}

To improve performance and training efficiency simultaneously, we select the dominant prototypes instead of the random prototypes. In DP-softmax we maintain a dominant queue for each class to store their similar classes, where the queue size $q$ is an important parameter that impacts both performance and training time. Table~\ref{tab-perf-hard-queue} shows the performances with different queue sizes and Fig.~\ref{fig:time-dominant-prototype} shows the corresponding training time. We can see that the performance increases as the queue size increases, but quickly saturates when $q$ reaches $100$ with only $3,000$ prototypes. Considering both performance and efficiency we set $q=100$ in our implementation. Compared with RP-softmax with $100,000$ prototypes, DP-softmax achieves better performance ($91.92\%$ vs. $90.30\%$ at FAR=$10^{-5}$) with much lower training time ($1.1$s vs. $1.6$s per iteration).

\begin{table}[!htb]
  \footnotesize \hspace{-2mm}
    \resizebox{\linewidth}{!}{%
    \begin{tabular}{|c||c|c|c|}
      \hline
      Method & FAR=$10^{-3}$ & FAR=$10^{-4}$ & FAR=$10^{-5}$ \\
      \hline
      DPS$_{10}$(0.3k)   & 96.62 & 92.48 & 85.12 \\
      DPS$_{20}$(0.6k)   & 96.77 & 93.69 & 86.57 \\
      DPS$_{50}$(1.5k)   & 97.16 & 94.37 & 88.29 \\
      DPS$_{100}$(3.0k)  & 97.70 & 96.17 & 91.92 \\
      DPS$_{300}$(10.0k) & 97.72 & 96.27 & 92.01 \\
      \hline
    \end{tabular}
    }
  \caption{The performances of DP-softmax (DPS) with different queue sizes on CASIA-IvS-Test, evaluated by VR(\%) at different FAR. DPS$_{q}(N_{iter})$ indicates the queue size is $q$ and the number of prototypes is $N_{iter}$. Note that there are two samples per class in the mini-batch, there are at most $Mq/2$ dominant prototypes.}
  \label{tab-perf-hard-queue}
\end{table}

In Table~\ref{tab-perf-queue-update}, we also compare the performances with and without queue updating, which demonstrates the effectiveness of queue updating.
\begin{table}[!htb]
  \footnotesize \hspace{-2mm}
    \resizebox{\linewidth}{!}{%
    \begin{tabular}{|c||c|c|c|}
      \hline
      Method & FAR=$10^{-3}$ & FAR=$10^{-4}$ & FAR=$10^{-5}$ \\
      \hline
      w/o update  & 97.54 & 95.58 & 90.77  \\
      update      & 97.70 & 96.17 & 91.92  \\
      \hline
    \end{tabular}
    }
  \caption{The performances of DP-softmax with and without queue updating, evaluated by VR(\%) at different FAR.}
  \label{tab-perf-queue-update}
\end{table}

\subsubsection{Softmax Formulation}\label{sec:exp-softmax-formulation}
Large-scale classification mainly involves a prototype selection strategy, which can be combined with any softmax formulation. Besides the traditional softmax~\cite{Sun-CVPR-2014}, the state-of-the-art A-softmax~\cite{liu2017sphereface} and AM-softmax~\cite{wang2018additive} can also be adopted. Table~\ref{tab-softmax-formulation} shows the results with different softmax formulations. We can see that A-softmax and AM-softmax have improved performance by introducing the margins and A-softmax has the best results.

\begin{table}[!htb]
  \footnotesize \hspace{-2mm}
    \resizebox{\linewidth}{!}{%
    \begin{tabular}{|c||c|c|c|}
      \hline
      Method & FAR=$10^{-3}$ & FAR=$10^{-4}$ & FAR=$10^{-5}$ \\
      \hline
      DPS+softmax     & 97.31 & 94.91 & 89.55 \\
      DPS+AM-softmax  & 97.60 & 95.59 & 90.73 \\
      DPS+A-softmax   & 97.70 & 96.17 & 91.92 \\
      \hline
    \end{tabular}
    }
  \caption{The performances of adopting different softmax formulations in large-scale classification, evaluated by VR(\%) at different FAR on CASIA-IvS-Test. The dominant prototype selection (DPS) is adopted.}
  \label{tab-softmax-formulation}
\end{table}

\subsection{Identity Volume}\label{sec-exp-examplar-selection}
\begin{figure}[!htb]
  \centering{}
  \includegraphics[width=0.5\textwidth]{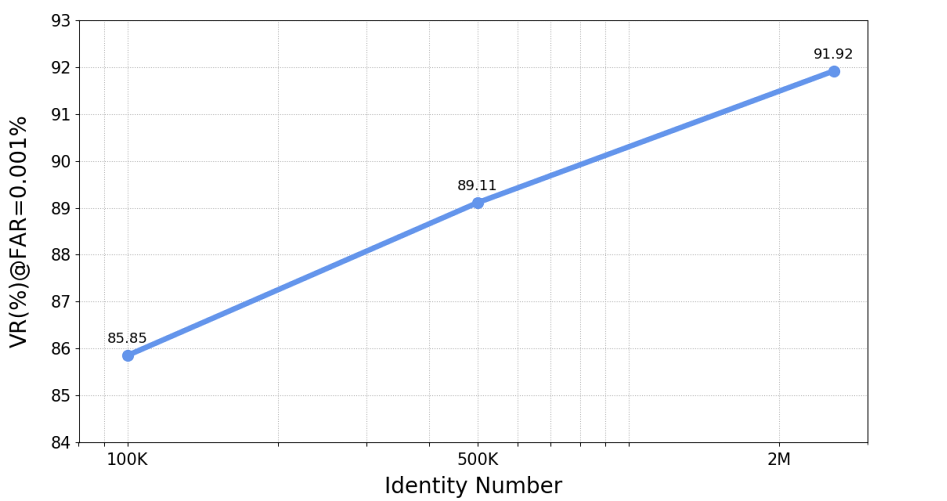}{}
  \caption{The performances under different identity volume, evaluated by VR(\%) at FAR=$10^{-5}$. }
  \label{fig-data-volume}
\end{figure}

It has been repeatedly observed that more data always delivers better performance~\cite{schroff2015facenet,sun2017revisiting}. Does the blessing of data still exist in IvS face recognition? To study this, we randomly sample a subset of $100$k, $500$k and $2$M identities from CASIA-IvS-Train and train the model, respectively. As shown in Fig.~\ref{fig-data-volume}, the performance grows logarithmically as identities increase, which is consistent with~\cite{sun2017revisiting}. We believe more identities provide more information about intra- and inter-variations, which delivers more discriminative features. Besides, it is suggested that the model can be further improved with more IvS data.

\subsection{Comparison Experiments}\label{sec-comparison-experiment}
In order to compare our method with the state of the arts, we choose several methods feasible on large-scale bisample data, including \textbf{Contrastive}~\cite{sun2014deep}, \textbf{Triplet}~\cite{schroff2015facenet}, \textbf{Lifted Struct}~\cite{oh2016deep}, \textbf{N-pairs}~\cite{sohn2016improved} and the Model A-D in Megaface challenge~\cite{nech2017level} (\textbf{MF-A} to \textbf{MF-D}). We also evaluate the large-scale classification methods in language models including Noise Contrastive Estimation (\textbf{NCE})~\cite{gutmann2010noise} and Hierarchical Softmax (\textbf{H-softmax})~\cite{bengio2003neural}. For fair comparison, all the methods adopt the ResNet64 architecture and their models are pretrained on Ms-Celeb-1M. In our implementation, for contrastive, each sample is paired with all the other ones in a mini-batch and the negative pairs are filtered by hard negative mining. For triplet, we adopt N-pairs batch construction~\cite{sohn2016improved} and anchor swapping~\cite{balntas2016learning} to construct the most triplets. Besides, online hard mining~\cite{schroff2015facenet} is performed to remove easy triplets. For N-pairs, we adopt the N-pair-mc loss to optimize each positive pair against all the related negative pairs and use the hard negative class mining to generate mini-batches with similar classes. For lifted struct, we directly use the released codes. For MF-A, we train the model with softmax on randomly selected $100,000$ classes. Then we finetune MF-A on the full data with the triplet loss as MF-B. For MF-C, we adopt the rotating softmax where $20,000$ random classes are selected in each epoch. Then we adopt the same triplet finetuning strategy to get MF-D. For NCE and H-softmax, directly training with the two losses cannot converge. In our implementation, we first train the models by the triplet loss and initialize the prototypes by the deep features as our LBL, making the training convergent.
For our methods, we first provide a naive baseline to perform softmax on IvS data named LBL(softmax), where in the final stage of CVC we train model only on $100,000$ classes (which are the most classes affordable by the machine) and the classes do not change as training proceeds. Besides, we report LBL with RP-softmax and DP-softmax.

\begin{figure*}[!htb]
  \centering
  \subfigure[CASIA-IvS-Test]{
  \label{fig-comp-caisa-ivs}
  \includegraphics[width=0.32\textwidth]{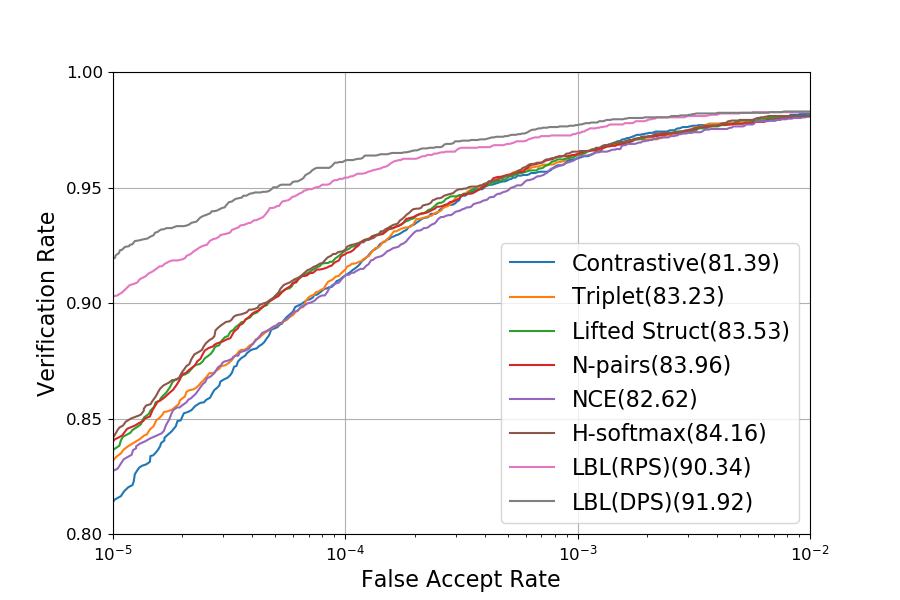}}
  \subfigure[Public-IvS]{
  \label{fig-comp-public-ivs}
  \includegraphics[width=0.32\textwidth]{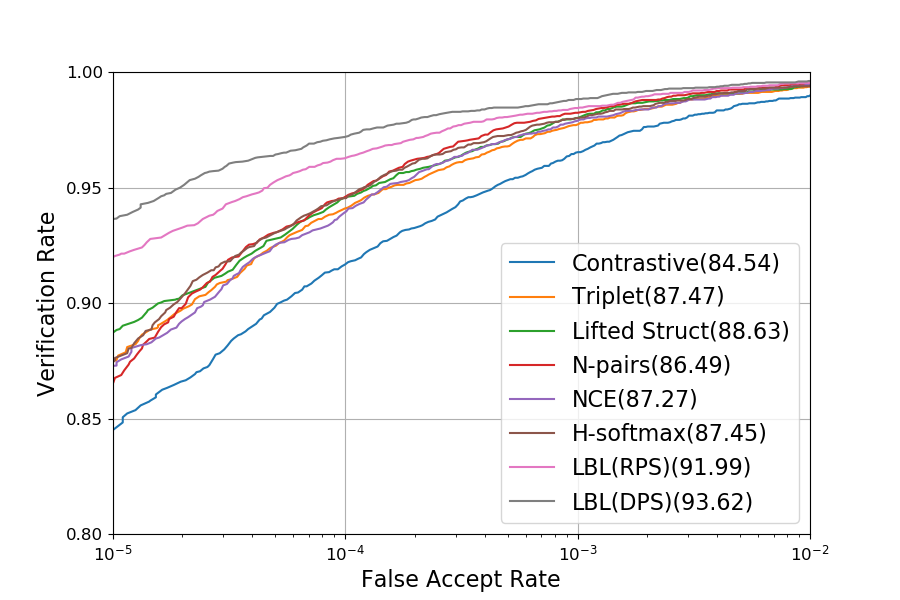}}
  \subfigure[Megaface-bisample]{
  \label{fig-comp-megaface-mini}
  \includegraphics[width=0.32\textwidth]{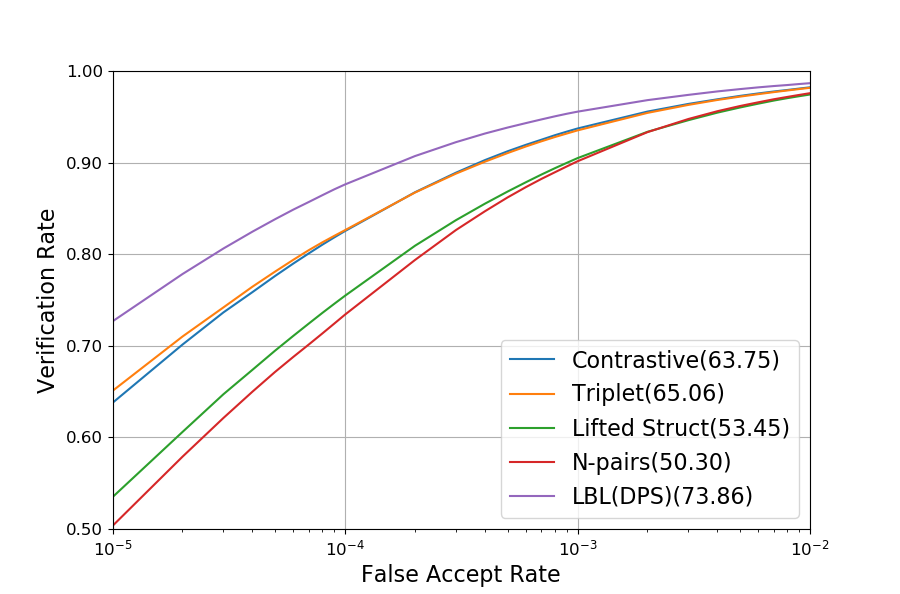}}
  \caption{Comparison of ROC curves on CASIA-IvS-Test, Public-IvS and Megaface-bisample. The values in the brackets are the VR(\%) at FAR=$10^{-5}$.}
  \label{fig-roc}
\end{figure*}

Table.~\ref{table-comp-ivs} shows the performances on the real-world CASIA-IvS-Test and the open Public-IvS. Fig.~\ref{fig-comp-caisa-ivs} and Fig.~\ref{fig-comp-public-ivs} show the corresponding ROC curves. During implementation, we find MF-A cannot achieve satisfactory performance since only a small part of data can be used. MF-C is hard to converge since the rotating softmax randomly initializes the prototypes periodically. After finetuned by the triplet loss on all the data, the models (MF-B and MF-D) still fail to get satisfactory performances due to the poor initializations.
As for our method LBL, we can see Public-IvS shows consistent results with CASIA-IvS-Test where our methods perform best. Besides, LBL significantly outperforms other methods on IvS data, especially at low FAR. The improvement at FAR=$10^{-5}$ is $84.16\%$ to $91.92\%$ on CASIA-IvS-Test and $88.63\%$ to $93.62\%$ on Public-IvS. The DP-softmax further improves the RP-softmax and achieves the best performance. LBL also achieves better recognition rates than the large scale classification methods in language models like NCE and H-softmax.

\begin{table*}[!htb]\small
 \tabcolsep 7pt
  \begin{center}
    \begin{tabular}{c||c|c|c||c|c|c}
      \hline
      {\multirow{2}{*}{Method}} & \multicolumn{3}{c||}{ CASIA-IvS-Test } & \multicolumn{3}{c}{ Public-IvS } \\
      \cline{2-7}
      {\multirow{2}{*}{}} & FAR=$10^{-3}$ & FAR=$10^{-4}$ & FAR=$10^{-5}$& FAR=$10^{-3}$ & FAR=$10^{-4}$ & FAR=$10^{-5}$ \\
      \hline
      Contrastive~\cite{sun2014deep}    & 96.25 & 91.17 & 81.39 & 96.52 & 91.71 & 84.54 \\
      \hline
      Triplet~\cite{schroff2015facenet}        & 96.41 & 91.39 & 83.23 & 97.72 & 94.11 & 87.47 \\
      \hline
      Lifted Struct~\cite{oh2016deep}   & 96.42 & 92.25 & 83.53 & 98.03 & 94.56 & \textbf{88.63} \\
      \hline
      N-pairs~\cite{sohn2016improved}    & 96.45 & 92.13 & 83.96 & \textbf{98.23} & \textbf{94.57} & 86.49 \\
      \hline
      NCE~\cite{gutmann2010noise}(Triplet-init) & 96.30 & 91.18 & 82.62 & 97.90 & 93.93 & 87.27 \\
      \hline
      H-softmax~\cite{bengio2003neural}(Triplet-init) & \textbf{96.50} & \textbf{92.36} & \textbf{84.16} & 98.01 & 94.54 & 87.45 \\
      \hline
      MF-A~\cite{nech2017level}    & 51.61 & 33.82 & 20.67 & 49.42 & 28.40 & 14.99 \\
      MF-B~\cite{nech2017level}    & 75.24 & 53.85 & 35.09 & 66.68 & 44.40 & 28.66 \\
      MF-C~\cite{nech2017level}    & 51.02 & 31.11 & 15.05 & 43.41 & 24.20 & 12.55 \\
      MF-D~\cite{nech2017level}    & 75.04 & 52.46 & 31.84 & 64.68 & 42.85 & 25.23 \\
      \hline
      \hline
      LBL(softmax)                 & 97.01 & 93.69 & 86.68 & 98.38 & 95.49 & 89.63\\
      LBL(RP-softmax)              & 97.43 & 95.40 & 90.34 & 98.44 & 96.29 & 91.99 \\
      LBL(DP-softmax)              & \textbf{97.70} & \textbf{96.17} & \textbf{91.92} & \textbf{98.83} & \textbf{97.21} &\textbf{93.62} \\
      \hline
    \end{tabular}
    \vspace{-2mm}
  \end{center}
  \caption{The performances of the state of the arts, evaluated by the VR(\%) at different FAR. The models are trained on CASIA-IvS-Train and evaluated on CASIA-IvS-Test and Public-IvS, with our method and the best baseline highlighted.}
  \label{table-comp-ivs}
\end{table*}

\subsection{Mimic Experiments on Megaface-bisample}\label{sec:exp-mimic}

To make our work reproducible, we mimic the large-scale bisample challenge on the open MF2~\cite{nech2017level} dataset and propose a new protocol Megaface-bisample. The MF2 contains $657,559$ identities which are much more than other datasets. We split MF2 into two subsets, MF2-thick and MF2-mini. The MF2-thick contains the identities having more than $15$ samples, which is used to simulate the well-posed dataset for pre-learning. The MF2-mini contains two randomly selected samples for each identity, which is used to simulate the bisample data. As for testing, we follow the BLUFR protocol~\cite{liao2014benchmark} on LFW~\cite{Huang-2007-LFW}. In summary, MF2-thick, MF2-mini and LFW-BLUFR simulate Ms-Celeb-1M, CASIA-IvS-Train and CASIA-IvS-Test, respectively. Specifically, MF2-thick has $46,000$ identities and $34.8$ samples per identity and MF2-mini has cleaned $649,790$ identities and $2$ samples per identity, whose image list will be released. As well known, MF2 has few celebrities and we have tried our best to ensure there is no identity overlap between MF2 and LFW. Although Megaface-bisample is not IvS data, it shares the same challenges: the weak intra-variations and model training scalability, as IvS data. Since there is no ID photo in MF2, we initialize the classification layer with avg-prototypes and construct the $\mathcal{NC}_{K}$ by avg-prototypes instead of ID features.

\begin{figure}[!htb]
  \centering{}
  \includegraphics[width=0.5\textwidth]{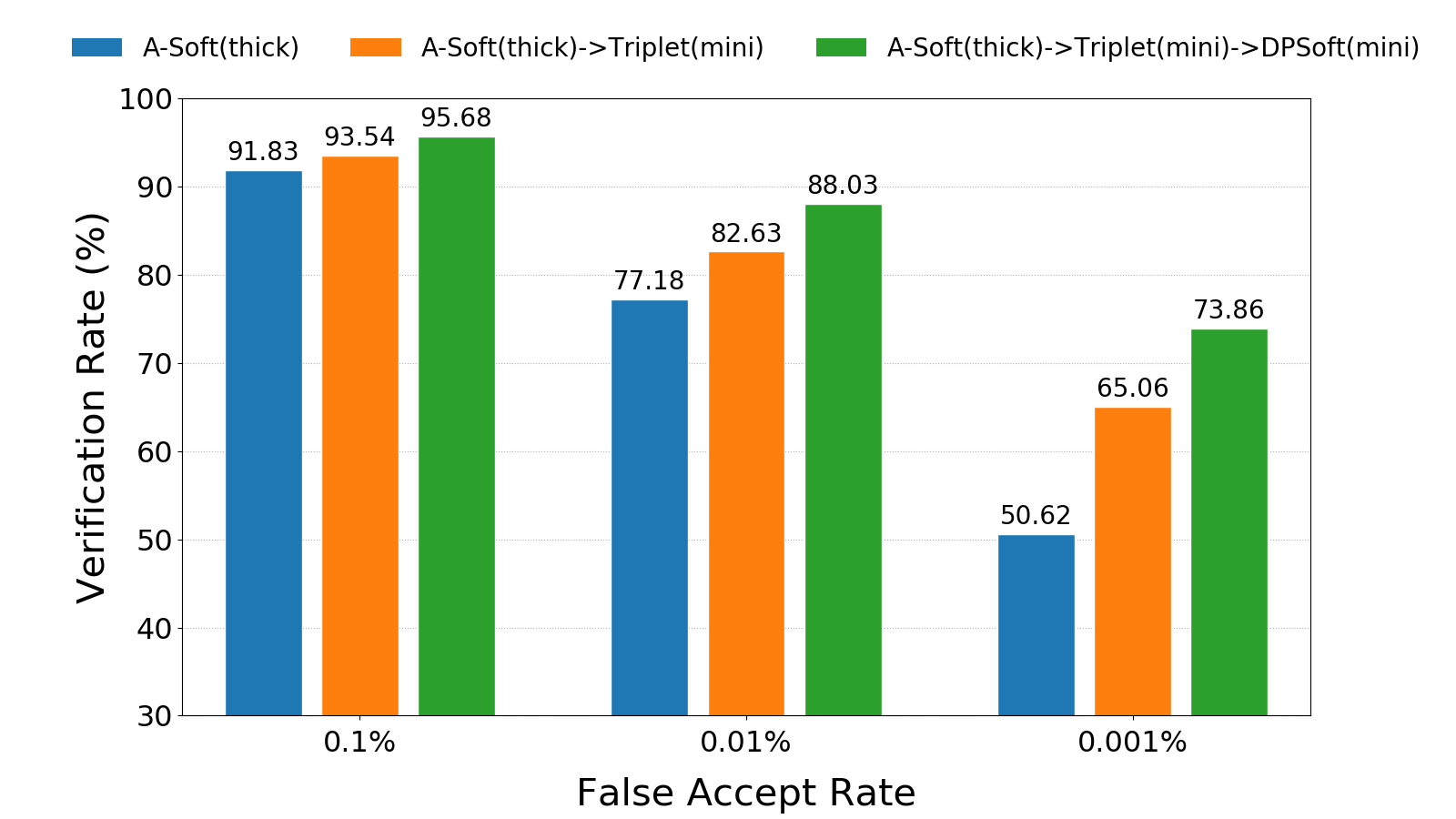}{}
  \caption{The intermediate results of CVC, following the Megaface-bisample protocol.}
  \label{fig:webface-mimic}
\end{figure}

First, to verify the effectiveness of the simulation, we re-implement the experiments of Table.~\ref{table-training-strategy} about the CVC training strategy. As shown in Fig.~\ref{fig:webface-mimic}, there is significant improvement after each stage. Besides, we try to train model from scratch on MF2-mini and find the training quickly falls into bad local optima. Since the results are consistent with the ones on CASIA-IvS, we believe Megaface-bisample can well simulate our task.

On Megaface-bisample we also compare our methods with the state of the arts in Table.~\ref{table:mimic-Comparative-BLUFR}, whose ROC curves are shown in Fig.~\ref{fig-comp-megaface-mini}. The proposed LBL still consistently outperforms the other methods and the improvement at FAR$=10^{-5}$ is over $8$ percent.

\begin{table}[!htb]

  \footnotesize \hspace{-2mm}
    \resizebox{\columnwidth}{!}{%
    \begin{tabular}{|c||c|c|c|}
      \hline
      {\multirow{2}{*}{Methods}} & \multicolumn{3}{c|}{ LFW-BLUFR } \\
      \cline{2-4}
       & FAR=$10^{-3}$ & FAR=$10^{-4}$ & FAR=$10^{-5}$  \\
      \hline
      Contrastive~\cite{sun2014deep}   & \textbf{93.74} & 82.53 & 63.75  \\
      \hline
      Triplet~\cite{schroff2015facenet}       & 93.54 & \textbf{82.63} & \textbf{65.06}  \\
      \hline
      Lifted Struct~\cite{oh2016deep}  & 90.50 & 75.46 & 53.45  \\
      \hline
      N-pairs~\cite{sohn2016improved}   & 90.16 & 73.40 & 50.30  \\
      \hline
      \hline
      LBL(DP-softmax)                          & \textbf{95.68} & \textbf{88.03} & \textbf{73.86}  \\
      \hline
    \end{tabular}
    }
  \caption{The verification rates, VR(\%), at different false acceptance rates (FAR) on LFW-BLUFR following the Megaface-bisample, with the top-2 results highlighted.}
  \label{table:mimic-Comparative-BLUFR}
\end{table}

\section{Conclusion}
This paper proposes a large-scale bisample learning (LBL) method to train deep neural networks on ID versus Spot (IvS) face data. Specifically, we develop a Classification-Verification-Classification (CVC) bisample training strategy that first transfers the knowledge from wild scenarios to IvS scenarios and then boosts the performance by large-scale classification. We also propose a dominant prototype softmax (DP-softmax) to perform $2$-million classification, which is used in the final stage of CVC. The DP-softmax diligently selects the dominant prototypes for each mini-batch, which improves the performance and reduces the training cost simultaneously. Experiments on a large real-world dataset show the proposed LBL significantly improves the IvS face recognition and the DP-softmax can perform effective classification with only $0.15\%$ of classes. Besides, we also release a Public-IvS dataset for open IvS evaluation and a new protocol Megaface-bisample to mimic the large-scale bisample learning task.

\section{Acknowledgments}
This work was supported by the Chinese National Natural Science Foundation Projects \#61876178, \#61806196, the National Key Research and Development Plan (Grant No.2016YFC0801002), and AuthenMetric R\&D Funds. Zhen Lei is the corresponding author.

\small
\bibliographystyle{IEEEtran}
\bibliography{reference_zxy}

\end{document}